\documentclass[10pt,twocolumn,letterpaper]{article}

\usepackage{cvpr}              

\usepackage[accsupp]{axessibility}  

\usepackage{graphicx}
\usepackage{amsmath}
\usepackage{amssymb}
\usepackage{booktabs}

\usepackage{enumitem}
\usepackage{multirow}
\usepackage{booktabs}
\newcommand{\tablestyle}[2]{\setlength{\tabcolsep}{#1}\renewcommand{\arraystretch}{#2}\centering\footnotesize}
\usepackage{amsfonts}
\usepackage{enumitem}
\usepackage{multirow}
\usepackage{booktabs}

\usepackage{xcolor}         

\usepackage{graphicx}
\usepackage{algorithmic}
\usepackage{algorithm}

\newcommand{\ieno}{\textit{i}.\textit{e}.}
\newcommand{\egno}{\textit{e}.\textit{g}.} 
\newcommand{\etcno}{\textit{etc}} 

\usepackage[capitalize]{cleveref}
\crefname{section}{Sec.}{Secs.}
\Crefname{section}{Section}{Sections}
\Crefname{table}{Table}{Tables}
\crefname{table}{Tab.}{Tabs.}

\def\cvprPaperID{6708} 
\def\confName{CVPR}
\def\confYear{2022}

\begin{document}

\title{Cloth-Changing Person Re-identification from A Single Image \\ with Gait Prediction and Regularization}

\author{
Xin Jin\textsuperscript{1,2}\thanks{This work was done when he was visiting Alibaba as a research intern.},
Tianyu He\textsuperscript{2},
Kecheng Zheng\textsuperscript{1},
Zhiheng Yin\textsuperscript{3},
Xu Shen\textsuperscript{2},
Zhen Huang\textsuperscript{1},
Ruoyu Feng\textsuperscript{1},\\
Jianqiang Huang\textsuperscript{2},
Zhibo Chen\textsuperscript{1}\thanks{Corresponding author.},
Xian-Sheng Hua\textsuperscript{2}\footnotemark[2]\\
\textsuperscript{1}University of Science and Technology of China, 
\textsuperscript{2}Alibaba Cloud Computing Ltd.\\
\textsuperscript{3}University of Michigan\\
{\tt\small \{jinxustc,zkcys001,hz13,ustcfry\}@mail.ustc.edu.cn,yzhiheng@umich.edu,chenzhibo@ustc.edu.cn}\\
{\tt\small \{timhe.hty,shenxu.sx,jianqiang.hjq,xiansheng.hxs\}@alibaba-inc.com}
}

\maketitle

\begin{abstract}

    Cloth-Changing person re-identification (CC-ReID) aims at matching the same person across different locations over a long-duration, \egno, over days, and therefore inevitably has cases of changing clothing. In this paper, we focus on handling well the CC-ReID problem under a more challenging setting, \ieno, just from a single image, which enables an efficient and latency-free person identity matching for surveillance. Specifically, we introduce \textbf{G}ait recognition as an auxiliary task to drive the \textbf{I}mage \textbf{ReID} model to learn cloth-agnostic representations by leveraging personal unique and cloth-independent gait information, we name this framework as \textbf{GI-ReID}. GI-ReID adopts a two-stream architecture that consists of an image ReID-Stream and an auxiliary gait recognition stream (Gait-Stream). The Gait-Stream, that is discarded in the inference for high efficiency, acts as a regulator to encourage the ReID-Stream to capture cloth-invariant biometric motion features during the training. To get temporal continuous motion cues from a single image, we design a Gait Sequence Prediction (GSP) module for Gait-Stream to enrich gait information. Finally, a semantics consistency constraint over two streams is enforced for effective knowledge regularization. Extensive experiments on multiple image-based Cloth-Changing ReID benchmarks, \egno, LTCC, PRCC, Real28, and VC-Clothes, demonstrate that GI-ReID performs favorably against the state-of-the-art methods.

\end{abstract}

\vspace{-5mm}
\section{Introduction}

\begin{figure}
  \centerline{\includegraphics[width=1.0\linewidth]{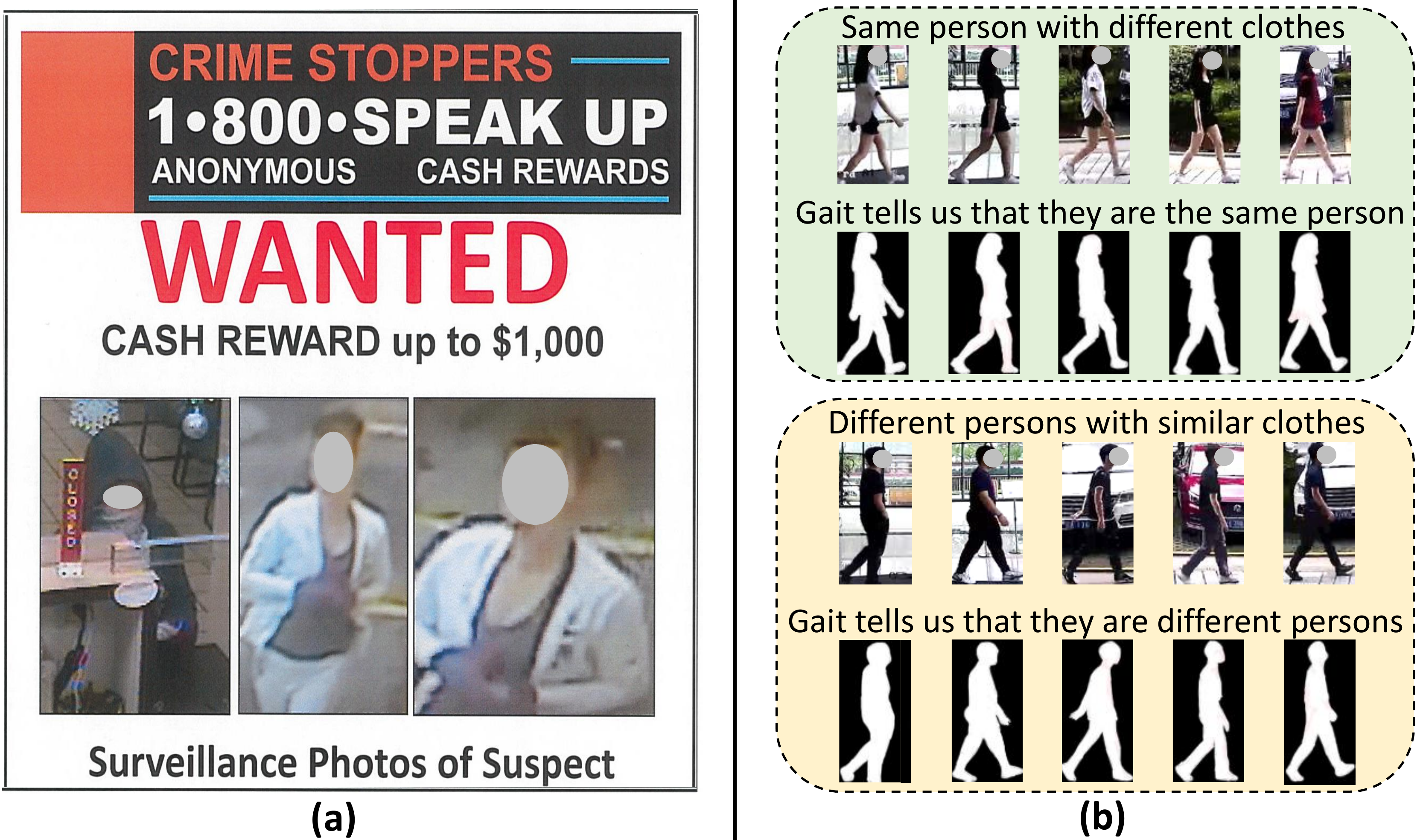}}
  \vspace{-3mm}
  \caption{(a) shows a realistic wanted case that a suspect changed her coat from black to white for hiding. (b) reveals that the gait of person could help ReID, especially when the identity matching meets the cloth-changing challenge (All faces in the images are masked for anonymization).}
  \vspace{-5mm}
  \label{fig:motivation}
\end{figure}

Person re-identification (ReID) aims at identifying a specific person across cameras, times, and locations. Abundant approaches have been proposed to address the challenging geometric misalignment among person images caused by diversities of human poses~\cite{su2017pose,zhao2017spindle,qian2018pose}, camera viewpoints~\cite{zhang2016learning,sun2019dissecting,jin2020uncertainty}, and style/scales~\cite{jin2020semantics,jin2020style}. These methods usually inadvertently assume that both query and gallery images of the same person have the \emph{same clothing}. In general, they perform well on the trained short-term datasets but suffer from signiﬁcant performance degradations when testing on a long-term collected ReID dataset~\cite{yang2019person,qian2020long,yu2020cocas,wan2020person}. Because large clothing variations occur over long-duration among these datasets, which seriously hinders the accuracy of ReID. For example, Figure~\ref{fig:motivation}(a) shows a realistic wanted case~\footnote{Information comes from https://www.wjr.com/2016/01/06/woman-wanted-in-southwest-detroit-bank-robbery/} where a suspect that captured by surveillance devices at different times/locations changed her coat from black to white, which makes ReID difficult, especially when she wears a mask and the captured images are of low quality.

In recent years, to handle the cloth-changing ReID (CC-ReID) problem, some studies have contributed some new datasets where clothing changes are commonplace (\egno, Celebrities-reID~\cite{huang2019beyond,huang2019celebrities}, PRCC~\cite{yang2019person}, LTCC~\cite{qian2020long}, Real28 and VC-Clothes~\cite{wan2020person}). They also propose some new algorithms that could learn cloth-agnostic representations for CC-ReID. For instance, Yang~\etal~\cite{yang2019person} propose a contour-sketch-based network to overcome the moderate cloth-changing problem. Similarly,  Qian~\etal~\cite{qian2020long}, Li~\etal~\cite{li2020learning}, and Hong~\etal~\cite{hong2021fine} all use body shape to tackle the CC-ReID problem. However, no matter of using a contour sketch or body shape, all these methods are prone to suffer from the estimation error problem. Because the single-view contour/shape inference (from 2D image) is extremely difficult due to the vast range of possible situations, especially when people wear thick clothes in winter. Besides, these contour-sketch-based or shape-based methods only focus on extracting \emph{static spatial cues} from persons as extra cloth-agnostic representations, the rich \emph{dynamic motion information} (\egno, gait, implied motion~\cite{kourtzi2000activation}) are often ignored.


In this paper, we explore to leverage the unique gait features that imply dynamic motion cues of a pedestrian to drive a model to learn cloth-agnostic and discriminative ReID representations. As shown in Figure~\ref{fig:motivation}(b), although it is hard to identify the same person when he/she wears different clothes, or to distinguish the different persons when they wear similar/same clothes, we can still leverage their unique/discriminative gaits to achieve correct identity matching. It is because that gait, as a unique biometric feature, has the superior invariance compared with other easy-changing appearance characteristics, \egno, face, body shape, contour~\cite{liu2015enhancing,zhang2018long}. Besides, gait can be authenticated at a long distance even with low-quality camera imaging.

Unfortunately, existing gait-related studies mainly rely on large video sequences~\cite{chao2019gaitset,Fan_2020_CVPR}. Capturing videos requires time latency and saving videos needs a large hardware storage cost, which are both undesirable for the real-time ReID applications. {Even the recent work~\cite{xugait2020gait} first attempts to achieve gait recognition from a single image, how to leverage gait feature to handle CC-ReID problem from a single image is still under-studied and this task is more challenging due to the potential viewpoint-variations and occlusions.}



\begin{figure*}
  \centerline{\includegraphics[width=0.99\linewidth]{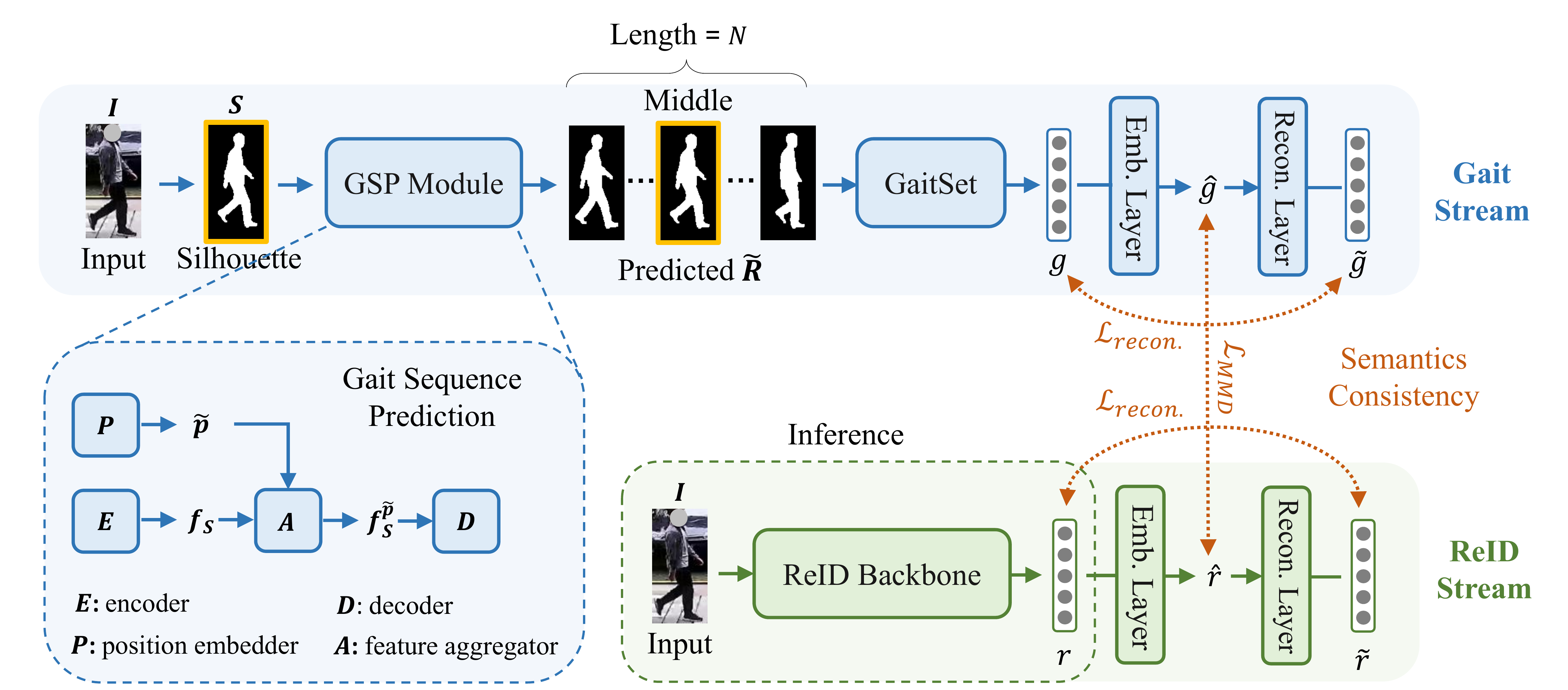}}
  \vspace{-1mm}
  \caption{Overview of the proposed GI-ReID, which consists of \emph{ReID-Stream} and \emph{Gait-Stream}, they are jointly trained with a high-level Semantics Consistency (SC) constraint. The \emph{Gait-Stream} plays the role of a regulator to drive \emph{ReID-Stream} to learn cloth-agnostic representations from a single image, and it is \textbf{discarded} in the inference for computational efficiency. Gait Sequence Prediction (GSP) module aims at predicting gait frames from an image. GaitSet~\cite{chao2019gaitset} is responsible for extracting discriminative gait features.}
  \label{fig:pipeline}
  \vspace{-5mm}
\end{figure*}

In this paper, we propose a \textbf{G}ait-assisted \textbf{I}mage-based \textbf{ReID} framework, termed as GI-ReID, which could learn cloth-agnostic ReID representations from a single image with the gait feature assistance. GI-ReID consists of a main image-based ReID-Stream and an auxiliary gait recognition stream (Gait-Stream). Figure~\ref{fig:pipeline} shows the entire framework. The Gait-Stream aims to regularize the ReID-Stream to learn cloth-agnostic features from a single RGB image for effective CC-ReID. It is discarded in the inference for the high efficiency. Since the comprehensive gait features extraction typically needs a gait video sequence as input~\cite{chao2019gaitset,Fan_2020_CVPR}, we introduce a new Gait Sequence Prediction (GSP) module for Gait-Stream to approximately forecast continuous gait frames from a single input query image, which enriches the learned gait information. Finally, to encourage the main ReID-Stream's efficient learning from Gait-Stream, we further enforce a high-level Semantics Consistency (SC) constraint for the same person over two streams' features. We summarize our main contributions as follows:




    
    

\begin{itemize}[leftmargin=*,noitemsep,nolistsep]
    \item We specially aimed at handling the challenging cloth-changing issue for image ReID to promote practical applications. A Gait-assisted Image-based cloth-changing ReID (GI-ReID) framework is proposed. As a regulator, the Gait-Stream in GI-ReID can be removed in the inference without sacrificing ReID performance. This reduces the dependency on the accuracy of gait recognition, making our method computationally efficient and robust.
    
    \item A well-designed Gait Sequence Prediction (GSP) module makes our method effective in the challenging image-based ReID scenarios. And, a high-level semantics consistency (SC) constraint enables an effective regularization over two streams, enhancing the distinguishing power of ReID-Stream under the cloth-changing setting.
    
\end{itemize}

With the gait prediction and regularization, GI-ReID achieves a state-of-the-art performance on the image-based cloth-changing ReID. It is also general enough to be compatible with the existing ReID-specific networks, except ResNet-50~\cite{he2016deep}, we
also use OSNet~\cite{zhou2019omni}, LTCC-shape~\cite{qian2020long}, and PRCC-contour~\cite{yang2019person} as our baselines for evaluation.

\vspace{-2mm}
\section{Related Work}

\subsection{Person Re-identification}

\noindent\textbf{General ReID.} Without cloth-changing cases, the general ReID has achieved a great success with the deep learning. It includes exploring fine-grained pedestrian feature descriptions~\cite{sun2018beyond,wang2018learning,fu2019horizontal,zhou2019omni}, and addressing spatial misalignment caused by (a) different camera viewpoints~\cite{sun2018dissecting,jin2020uncertainty}, (b) different poses~\cite{su2017pose,ge2018fd,qian2018pose}, (c) semantics inconsistency~\cite{zhang2019DSA,jin2020semantics}, (d) occlusion/partial-observation~\cite{zhuo2018occluded,miao2019pose,zheng2015partial,he2018deep}, \etcno. These methods rely substantially on static spatial texture information. However, when person ReID meets changing clothes, the texture information is not so reliable since it changes significantly even for the same person. Compared to static texture, the gait information, as a discriminative biometric modality, is more consistent and reliable.




\noindent\textbf{Cloth-Changing ReID.} Considering the wider application range and greater practical value of Cloth-Changing ReID (CC-ReID), more and more studies pay their attention to solve this challenging problem. Huang~\etal~\cite{huang2019beyond,huang2019celebrities} propose to use vector-neuron capsules~\cite{sabour2017dynamic} to perceive cloth changes of the same person. Yang~\etal~\cite{yang2019person}, Qian~\etal~\cite{qian2020long}/ Li~\etal~\cite{li2020learning}, Yu~\etal~\cite{yu2020cocas}/Wan~\etal~\cite{wan2020person} propose to leverage contour sketch, body shape, face/hairstyle to assist ReID under the cloth-changing setting, respectively. Nevertheless, these methods usually suffer from estimation error due to the difficulty of obtaining external cues (\egno, body shape, face, \etcno). Besides, they also ignore the exploration of discriminative dynamic motion cues, like gait.





FITD~\cite{zhang2018long} solves the cloth-changing ReID problem based on true motion cues {of videos}. Our work differs from FITD for at least three perspectives: 1). FITD uses motion information derived from dense trajectories (optical flow), which requires continuous \textbf{video sequences}. Our GI-ReID handles cloth-changing ReID from \textbf{a single image} with gait prediction and regularization, which is more challenging and practical. 2). FITD directly uses human motion cues to complete ReID, which relies on the accurate motion prediction and may suffer from estimation errors. Our GI-ReID just takes the gait recognition task as a regulator to drive the main ReID model to learn cloth-independent features, which makes our method less sensitive to gait estimation errors. 3). FITD only characterizes temporal motion patterns for ReID, ignoring other distinguishable local spatial cues, like personal belongings (\egno, backpacks). Our GI-ReID not only explores dynamic gait cues, but also learns from raw RGB images, leading more comprehensive features.

\begin{table}
\vspace{1mm}
\scriptsize
\caption{Differences between Gait Recognition and CC-ReID.}
\vspace{-2mm}
\setlength{\tabcolsep}{0.5mm}{
\begin{tabular}{|c|c|c|}
\hline
Task            & Gait Recognition            & Cloth-Changing Person ReID                                                                                             \\ \hline
Data format &
  \begin{tabular}[c]{@{}c@{}}Gait Energy Image (GEI) / \\ Sequence set of silhouette / \\ Video sequences\end{tabular} &
  \begin{tabular}[c]{@{}c@{}}\textbf{Discontinuous} RGB images \\ across cameras\end{tabular} \\ \hline
{Datasets}    & \begin{tabular}[c]{@{}c@{}}USF, CASIA-B, \\ OU-ISIR, OU-MVLP, \etcno.\end{tabular} & \begin{tabular}[c]{@{}c@{}}COCAS, PRCC, LTCC, \\ Real28, VC-Clothes, \etcno.\end{tabular}                            \\ \hline
\begin{tabular}[c]{@{}c@{}}Unsolved \\ problems \end{tabular} &
  \begin{tabular}[c]{@{}c@{}}1) Viewing angles (\egno, frontal view); \\ 2) Occlusion, body incompleteness; \\ 3) Cluttered/complex background;\end{tabular} &
  Clothes variation \\ \hline
\end{tabular}}
\vspace{-6mm}
\label{dff_comparison}%
\end{table}

\vspace{-1mm}
\subsection{Gait Recognition and Prediction}
\vspace{-1mm}
\noindent\textbf{Gait recognition}~\cite{muramatsu2014gait,liu2015enhancing,makihara2017joint,chao2019gaitset,carley2019person,li2020gait,Fan_2020_CVPR,elharrouss2020gait,xugait2020gait} directly uses gait sequence for identity matching, which is also cloth-independent, but different from our work and cannot be directly applied into {image-based cloth-changing ReID}. We clarify the differences between two tasks in detail in Table~\ref{dff_comparison}: this paper focuses on image-based cloth-changing ReID where large viewpoint variations, occlusion, and complex environments make gait recognition failed. And, these gait sequence based methods are not optimal for the image-based CC-ReID. Thus, we just take the gait recognition as an auxiliary regularization to drive ReID model to learn cloth-agnostic representations, which makes our method robust to the recognition errors. Moreover, the gait representations can be grouped into model-based~\cite{nixon2009model,liao2020model,li2020end} and appearance-based~\cite{carley2019person,chao2019gaitset,Fan_2020_CVPR}. The first one relies on human pose, while the latter relies on silhouettes. We use silhouette as gait representation for simplicity and robustness.


\noindent\textbf{Gait Prediction} from a single frame, or said, the field of video frame prediction (\ieno, motion prediction) has been widely studied and achieved a
great success~\cite{guen2020disentangling,hsieh2018learning,liu2019deep,niklaus2018context,xugait2020gait}, which verifies the feasibility of our work. This task is very challenging, that’s why {we carefully design the gait sequence prediction module while indirectly using the prediction results} in a robust regularization manner to help cloth-changing ReID.

\vspace{-1mm}
\section{Proposed GI-ReID Framework}



GI-ReID framework aims to fully exploit the unique human gait to handle the cloth-changing challenge of ReID just depending on a single image. Figure~\ref{fig:pipeline} shows the flowchart of the entire framework. Given a single person image, its silhouette (\ieno, mask) will be first extracted as input to the Gait-Stream using semantic segmentation
methods, such as PointRend~\cite{kirillov2020pointrend}. With the proposed gait sequence prediction (GSP) module, we could predict a gait sequence with more comprehensive gait information, which is then fed into the subsequent recognition network (GaitSet~\cite{chao2019gaitset}) to extract discriminative gait features. Through a high-level semantics consistency (SC) constraint, the cloth-independent Gait-Stream acts as a regulator to encourage the main ReID-Stream to capture cloth-agnostic features from a single RGB image. We discuss the details of each component in the following sections.

\subsection{The Auxiliary Gait-Stream}\label{sec:gait_stream}

Gait-Stream is composed of two parts: Gait Sequence Prediction (GSP) module and the pre-trained gait recognition network (GaitSet~\cite{chao2019gaitset}). GSP is designed for gait information augmentation. Then, GaitSet extracts cloth-independent and discriminative motion feature cues from augmented gait to guide/regularize ReID-Stream's training.


\textbf{Gait Sequence Prediction (GSP) Module:}~\label{sec:GSP}
GSP module aims to predict a gait sequence that contains continuous gait frames. This module is related to the general video frame prediction task (\ieno, frame interpolation and extrapolation studies~\cite{niklaus2018context,hsieh2018learning,liu2019deep,guen2020disentangling}), and gait sequence prediction can be deemed as a ``gait frame synthesis'' process.


As shown in Figure~\ref{fig:pipeline}, GSP is based on an auto-encoder architecture~\cite{doersch2016tutorial} with feature encoder $E$ and decoder $D$. In order to reduce the prediction ambiguity and difficulty (\egno, given a dangling arm, it is hard to guess whether it will rise or fall in the next frame), we manually integrate an extra prior information of \textbf{middle frame index} into the inner learned feature through a position embedder $P$ and a feature aggregator $A$. Intuitively, the \textbf{middle frame index} means that the input gait silhouette corresponds to the middle result of the predicted gait sequence. Such prior knowledge aims to drive GSP module to predict the adjacent walking statuses \emph{before and after} the current input walking status so as to reduce prediction ambiguity.







\noindent\textbf{(1). Encoder.} Given a silhouette input $S$, the encoder $E$ aims to extract a dimension-shrinked compact feature:
\vspace{-1mm}
\begin{equation}
    \begin{aligned}
        f_S = E(S).
    \end{aligned}
\end{equation}
Specific/detailed network structures (including other components in Gait-Stream) can be found in \textbf{Supplementary}.

\noindent\textbf{(2). Position Embedder and Feature Aggregator.\label{sec:middle}} Considering the prediction ambiguity~\cite{meyer2015phase}, we introduce a \emph{middle frame input principle}, which assumes that the input silhouette always corresponds to the \textbf{middle} one of the predicted gait sequence. During the GSP training, we take the gait frame in the middle position of the ground truth gait sequence as input to GSP, and use a one-dimensional vector $p \in \mathbb{R}^1$, to denote such \emph{position label}. Given a ground truth gait sequence with $N$ frames, the position label $p_{mid} \in \mathbb{R}^1$ of the input middle gait is defined as $p_{mid} = N // 2$ which indicates the relative position relationship of input frame to the entire sequence. For convenience, we convert position label to one-hot vector to calculate loss. In formula, the position embedder $P$ works as:
\vspace{-1mm}
\begin{equation}
    \begin{aligned}
        \widetilde{p} = P(S) \hspace{0.5mm} \in \hspace{0.5mm} \mathbb{R}^1, \hspace{3mm} \mathcal{L}_{position} = ||\widetilde{p}-p_{mid}||_2^2,
    \end{aligned}
    \label{eq:loss_position}
\end{equation}
where we compare the embedded position output $\widetilde{p}$ with the ground truth $p_{mid}$ to construct a \emph{position} loss $\mathcal{L}_{position}$. $P$ is to build a mapping between input and middle position.

Feature aggregator $A$, implemented by a fully connected layer, is inserted between the encoder and the decoder to convert the raw encoded features $f_{S}$ into \emph{middle-position-aware} features $f_{S}^{\widetilde{p}}$ by taking the embedded middle position information $\widetilde{p}$ into account for the following decoder, which explicitly tells the decoder that we need to predict the gait statuses \emph{before and after} the current input middle gait status, and thus reduces prediction ambiguity for the predicted results. This feature aggregation process is formulated as:
\begin{equation}
    \begin{aligned}
        f_S^{\widetilde{p}} = A([\hspace{0.5mm} f_S,\hspace{0.5mm} \widetilde{p} \hspace{0.5mm} ]),
    \end{aligned}
    \label{eq:aggregation}
\end{equation}
where $[\cdot]$ means a simple concatenation.


\noindent\textbf{(3). Decoder.} We feed the aggregated feature $f_S^{\widetilde{p}}$ into the decoder $D$, which has a symmetrical structure to that of the encoder $E$, to predict the gait sequence with a pre-defined fixed number of frames $N$. Such process is formulated as,
\begin{equation}
    \begin{aligned}
        \widetilde{R} = D(f_S^{\widetilde{p}}) \hspace{0.5mm} \in \hspace{0.5mm} \mathbb{R}^{N*h*w}, \hspace{3mm} \mathcal{L}_{pred.} = ||\widetilde{R} - GT||_2^2,
    \end{aligned}
    \label{eq:loss_pred}
\end{equation}
where $(h, w)$ denotes the $(height, width)$ of predicted gait frames, same as the input silhouette image. A prediction loss $\mathcal{L}_{pred.}$ is calculated to ensure the predicted gait sequence results is consistent with the ground truth (GT).



\textbf{Gait Feature Extraction:}
The predicted gait sequence $\widetilde{R}$ is fed into the pre-trained GaitSet~\cite{chao2019gaitset} to learn discriminative and cloth-independent gait feature $g$. GaitSet is a set-based gait recognition model that takes a set of silhouettes as an input and aggregates features over frames into a set-level feature, which is formulated as $g = GaitSet(\widetilde{R})$. More details are presented in \textbf{Supplementary}.

\subsection{The Main ReID-Stream}



The backbone of the ReID-Stream could be any of the off-the-shelf networks, such as commonly-used ResNet-50~\cite{he2016deep}, ReID-specific PCB~\cite{sun2018beyond}, MGN~\cite{wang2018learning}, and OSNet~\cite{zhou2019omni}. And, we use the widely-adopted classification loss~\cite{sun2018beyond,fu2019horizontal}, and triplet loss with batch hard mining \cite{hermans2017defense}) on the ReID feature vector $r$ as basic optimization objectives for training. The feature $r$ is finally used for reference.

\subsection{Joint Learning of Two Streams}

Due to the potential rough silhouette extraction and the gait sequence prediction errors of GSP module, it is very difficult to directly exploit the gait information alone to complete effective ReID. Experimentally, we have attempted to conduct CC-ReID with only the predicted gait sequence $\widetilde{R}$ as input, and found this scheme failed to deliver good results (see ablation study for more details). Therefore, to exploit the cloth-independent merits of the gait information while avoiding the above-mentioned issues, we propose to jointly train Gait-Stream and ReID-Stream through a high-level semantics consistency (SC) constraint, where gait characteristics is taken as a regulator to drive the cloth-agnostic feature learning of ReID-Stream. Note that the SC constraint is also not needed in the inference.



\noindent\textbf{Semantics Consistency (SC) Constraint.} SC constraint is essentially related to the common feature learning works, such as knowledge distillation~\cite{hinton2015distilling}, mutual learning~\cite{zhang2018deep}, and knowledge amalgamation~\cite{ye2019student}. Our SC constraint differs from them mainly in two perspectives: 1). SC is to encourage a high-level common feature learning from two modalities (dynamic gait and static RGB image). 2). SC ensures information integrity for each stream/modality.

The details of the SC constraint are shown in Figure~\ref{fig:pipeline}. The learned gait feature $g$ of Gait-Stream and ReID feature $r$ of ReID-Stream are first transformed to a common and interactable space, via an embedding layer: $\hat{r} = Emb.(r)$ and $\hat{g} = Emb.(g)$, where $\hat{r}$ and $\hat{g}$ have the same feature dimensions. Then, we enforce the transformed features $\hat{r}$ and $\hat{g}$ to be closed to each other by minimizing the Maximum Mean Discrepancy (MMD)~\cite{gretton2012kernel}. MMD is a distance metric to measure the domain mismatch for probability distributions. We use it to measure the high-level semantics discrepancy between the transformed features $\hat{r}$ and $\hat{g}$, and minimize it to drive ReID-Stream to pay more attention to cloth-independent gait biometric. An empirical approximation to the MMD distance of $\hat{r}$ and $\hat{g}$ is simplified as follows:
\begin{equation}
    \begin{aligned}
        \mathcal{L}_{MMD} = \|{\mu}(\hat{g}) - {\mu}(\hat{r}) \|_2^2 +  \|{\sigma}(\hat{g}) - {\sigma}(\hat{r}) \|_2^2,
    \end{aligned}
    \label{eq:loss_MMD}
\end{equation}
where $\mu(\cdot), \sigma(\cdot)$ denotes the mean, variance calculation functions for the transformed features $\hat{r}$ and $\hat{g}$. 



To avoid the information lost caused by feature regularization with SC constraint, we further enforce a reconstruction penalty to ensure that the transformed features $\hat{g}$ and $\hat{r}$ could be recovered to original versions. Specifically, we reconstruct the original output features through a \emph{Recon.} layer (implemented by FC layer): $\widetilde{r} = Recon.(\hat{r})$ and $\widetilde{g} = Recon.(\hat{g})$, and calculate the corresponding reconstruction loss as follows:
\begin{equation}
    \begin{aligned}
        \mathcal{L}_{recon.} = \|\widetilde{g} - g \|_2^2 +  \|\widetilde{r} - r \|_2^2.
    \end{aligned}
    \label{eq:loss_recon}
\end{equation}

\noindent\textbf{Training Pipeline.} The whole training process of the proposed GI-ReID consists of three stages: 1). Pre-training GaitSet~\cite{chao2019gaitset} for gait feature extraction. 2). Joint Training for the proposed gait sequence prediction (GSP) module and GaitSet in Gait-Stream on gait-related datasets. 3). Joint Training for Gait-Stream and ReID-Stream on CC-ReID-related datasets. More details are provided in \textbf{Supplementary}, including pseudo code and loss balance strategy.


\section{Experiment}

\subsection{Datasets, Metric and Experimental Setups}
\noindent\textbf{Datasets Details.} We use four recent cloth-changing ReID datasets Real28~\cite{wan2020person}, VC-Clothes~\cite{wan2020person}, LTCC~\cite{qian2020long}, PRCC~\cite{yang2019person}, and one general video ReID dataset MARS~\cite{zheng2016mars} (to highlight the difficulty and necessity of image-based CC-ReID) to perform experiments. Table~\ref{tab:datasets} gives a brief information and comparison of these ReID datasets. More detailed introductions can be found in \textbf{Supplementary}.

\begin{table}[htbp]
  \footnotesize
  \vspace{-2mm}
  \centering
  \caption{Brief introduction and comparison of datasets.}
  \vspace{-3mm}
  \setlength{\tabcolsep}{0.6mm}{ 
    \begin{tabular}{c|c|c|c|c|c}
    \toprule
          & MARS  & Real28 & VC-Clothes & LTCC & PRCC \\
    \hline
    Category & Video & Image & Image & Image & Image\\
    Photo Style & Real  & Real  & Synthetic & Real & Real \\
    Scale & Large  & Small  & Large & Large & Large \\    
    Cloth Change & No    & Yes   & Yes   & Yes & Yes \\
    Identities & 1,261 & 28    & 512   & 152 & 221 \\
    Samples & 20,715 & 4,324 &  19,060 & 17,138 & 33,698\\
    Cameras & 6     & 4     & 4     & N/A & 3\\
    Usage & Train\&Test & Test  & Train\&Test & Train\&Test & Train\&Test\\
    \bottomrule
    \end{tabular}}%
    \vspace{-3mm}
  \label{tab:datasets}%
\end{table}%

\noindent\textbf{Evaluation Metrics.} We use the cumulative matching characteristics (CMC) at Rank-1/-10/-20,  and mean average precision (mAP) to evaluate the performance.


\noindent\textbf{Experimental Setups.} We build \textbf{three} kinds of different experiment settings to comprehensively validate the effectiveness of gait biometric for person ReID, and also validate the rationality/superiority of the proposed gait prediction and regularization in our GI-ReID framework: (1) Real Cloth-Changing Image ReID, (2) General Video ReID, and (3) Imitated Cloth-Changing Video ReID. In the main manuscripts, to save space and highlight core contributions of our paper, we only present the results related to the most challenging setting of (1) Real Cloth-Changing Image ReID. The rest results about (2)(3) are in \textbf{Supplementary}.

For (1) Real Cloth-Changing Image ReID, we employ real image-based cloth-changing datasets Real28~\cite{wan2020person}, VC-Clothes~\cite{wan2020person}, LTCC~\cite{qian2020long}, and PRCC~\cite{yang2019person} for experiments to validate the effectiveness of GSP module, SC constraint, and also compare our GI-ReID with SOTA cloth-changing ReID methods. In this setting, GSP module and GaitSet are both first pre-trained on gait-specific datasets CASIA-B~\cite{chao2019gaitset} and then fine-tuned on the CC-ReID datasets with the SC constraint $\mathcal{L}_{MMD}\&\mathcal{L}_{recon.}$ and the ReID supervisions. ResNet-50~\cite{he2016deep}, OSNet~\cite{zhou2019omni}, LTCC-shape~\cite{qian2020long}, and PRCC-contour~\cite{yang2019person} are taken as ReID backbone for comparisons.

\subsection{Ablation Study} \label{sec:AB}

Baseline means the model that only ingests RGB images.

\noindent\textbf{Results of Real Cloth-Changing Image ReID.} We conduct ablation experiments on three cloth-changing datasets Real28, VC-Clothes, and LTCC. Real28 is too small for training, so we train model on VC-Clothes and only test on Real28~\cite{wan2020person}. In Table~\ref{tab:CImgReID}, we see that 1) All Gait-Stream (GS) related schemes achieve obvious gains (over 2.7\% in mAP) over \emph{Baseline}, which demonstrates the effectiveness of using gait to handle cloth-changing issue. 2) With the well-designed GSP module, \emph{Baseline+ GS-GSP (concat)} outperforms the ablated scheme \emph{Baseline+GS (concat)} by 3.3\%/6.7\%/2.3\% in mAP on Real28/VC-Clothes/LTCC, which demonstrates the effectiveness of gait sequence prediction (GSP) on gait information augmentation. Note that \emph{Baseline+GS (concat)} just uses Gait-Stream (GS) but removes GSP, where we duplicate the only available single person silhouette as input to GaitSet. 3) Semantics consistency (SC) performs well in the cloth-changing settings, it helps our scheme GI-ReID achieve the best performance on the most evaluation cases while saving computational cost by discarding Gait-Stream in the inference.

\begin{table}
  \centering
  \footnotesize
  \caption{Performance (\%) comparison on the real image-based cloth-changing datasets Real28, VC-Clothes, LTCC. {GS-GSP} means Gait-Stream (GS) with gait sequence prediction (GSP) module. The ReID backbone is ResNet-50. `Standard' is the setting where the images in the test set with the same identity and camera view are discarded when computing mAP/Rank-1~\cite{qian2020long}.}
  \vspace{-2mm}
  \setlength{\tabcolsep}{0.7mm}{
    \begin{tabular}{ccc|cc|cc}
    \toprule
    \multirow{2}[4]{*}{Methods} & \multicolumn{2}{c}{Real28} & \multicolumn{2}{|c}{VC-Clothes} & \multicolumn{2}{|c}{LTCC (Standard)} \\
\cmidrule{2-7}          & mAP   & Rank-1 & mAP   & Rank-1 & mAP   & Rank-1 \\
    \midrule
    Baseline & 4.1  & 6.7  & 49.1  & 53.7  & 23.2   & 55.1 \\
     + GS (concat) & 6.8  & 7.9  & 52.3  & 58.9  & 26.5  & 60.0 \\
     + GS-GSP (concat) & 10.1  & 10.8  & \textbf{59.0}  & 63.7  & 28.8  & \textbf{64.5} \\
     + GS-GSP + SC (ours) & \textbf{10.4}  & \textbf{11.1}  & 57.8  & \textbf{64.5}  & \textbf{29.4}  & 63.2 \\
    \bottomrule
    \end{tabular}}%
    \vspace{-4mm}
  \label{tab:CImgReID}%
\end{table}%

\begin{table}[htp]
  \centering
  \footnotesize
  \caption{Performance (\%) comparison on the cloth-changing dataset LTCC. Such experiment aims to show that our GI-ReID can bring gains because of the exploration of gait information, rather than simply introducing silhouettes (\ieno, human masks). The ReID backbone is ResNet-50.}
  \vspace{-3mm}
  \setlength{\tabcolsep}{8mm}{
    \begin{tabular}{c|c|c}
    \hline
    \multirow{2}[1]{*}{Methods} & \multicolumn{2}{c}{LTCC (Cloth-Changing)} \\
\cline{2-3}          & mAP   & Rank-1 \\
    \hline
    Baseline & 8.10 & 19.58  \\
    Silhouette-ReID & 7.04 & 17.92     \\
    GI-ReID (ours) & \bf 10.38 & \bf 23.72  \\
    \hline
    \end{tabular}}%
  \vspace{-2mm}
  \label{tab:maskreid}%
\end{table}%




\noindent\textbf{Improvement Comes
From Gait Prediction, Not Silhouettes Usage.} We believe that our GI-ReID could successfully address the cloth-changing ReID problem from a single image is indeed because it effectively leverages the gait prediction, instead of the introduction of human silhouettes (\ieno, masks). To prove that, we additionally design a scheme of \emph{Silhouette-ReID} that directly takes the person RGB-Silhouette pair as input to ReID model (following~\cite{song2018mask,chen2018person}), and compare it with our GI-ReID on the cloth-changing ReID dataset LTCC. ResNet-50 is taken as ReID backbone for all schemes for comparison fairness. As shown in Table~\ref{tab:maskreid}, we found that \emph{Silhouette-ReID} is even inferior to the baseline scheme \emph{Baseline (ResNet-50)} by 1.06\% in mAP under the cloth-changing setting. We analyze that directly using silhouette to remove the background clutters in pixel-level will make ReID model pay more attention on the foreground objects' appearance/clothes color information, which is unexpected and unreliable for cloth-changing ReID, and thus leads to a performance drop.


\noindent\textbf{Study on Directly Using Gait Recognition Methods for Cloth-Changing ReID.} As we have discussed in the related work, directly using the algorithms of gait recognition for solving cloth-changing ReID problem is not optimal, especially in the image-based CC-ReID scenarios. Experimentally, we compare the proposed GI-ReID with two popular pure gait recognition works, GaitSet~\cite{chao2019gaitset} and PA-GCR~\cite{xugait2020gait}. GaitSet needs a set/sequence of person silhouettes as input, but recently-released cloth-changing ReID datasets are image datasets that lack of continuous frames for the same person. Thus, we duplicate the only available single one person silhouette to a set as input to approximately apply GaitSet into image-based CC-ReID task. As shown in Table~\ref{tab:gaitreid}, these pure gait recognition works of GaitSet~\cite{chao2019gaitset} and PA-GCR~\cite{xugait2020gait} are both inferior to the baseline scheme \emph{Baseline (ResNet-50)} in mAP under the cloth-changing setting, which indicates that simply using gait biometric for person matching can not work well for cloth-changing ReID, our gait prediction and regularization idea performs better for handling CC-ReID, especially for the image-based CC-ReID.

\vspace{-.5mm}
\begin{table}[htp]
  \centering
  \footnotesize
  \caption{Performance (\%) comparison on the cloth-changing dataset LTCC. Such experiment aims to show that these pure gait recognition works can not work well for cloth-changing ReID. The ReID backbone is ResNet-50.}
  \vspace{-2.5mm}
  \setlength{\tabcolsep}{8.5mm}{
    \begin{tabular}{c|c|c}
    \hline
    \multirow{2}[1]{*}{Methods} & \multicolumn{2}{c}{LTCC (Cloth-Changing)} \\
\cline{2-3}          & mAP   & Rank-1 \\
    \hline
    Baseline & 8.10 & 19.58  \\
    GaitSet~\cite{chao2019gaitset} & 2.14 & 7.22     \\
    PA-GCR~\cite{xugait2020gait} & 3.36 & 9.01     \\
    GI-ReID (ours) & \bf 10.38 & \bf 23.72  \\
    \hline
    \end{tabular}}%
    \vspace{-4.mm}
  \label{tab:gaitreid}%
\end{table}%

\begin{table*}[t]\centering\vspace{-2mm}
    \caption{Study on the different design choices in the (a)(b) GSP module, and (c) SC constraint of our GI-ReID framework. `Cloth-Changing' setting means that the images with same identity, camera view and clothes are discarded during the testing.}
    \vspace{-3mm}
	\captionsetup[subffloat]{justification=centering}
	\subfloat[Study on the gait prediction length $N$.\label{tab:GSP_len}]{
		\tablestyle{3pt}{.7}
            \begin{tabular}{ccccc}
            \toprule            \multirow{3}[6]{*}{Methods} & \multicolumn{4}{c}{LTCC} \\
        \cmidrule{2-5}          & \multicolumn{2}{c}{Standard} & \multicolumn{2}{c}{Cloth-Changing} \\
        \cmidrule{2-5}          & mAP   & Rank-1 & mAP   & Rank-1 \\
            \midrule
            Baseline & 23.2  & 55.1  & 8.1   & 19.6 \\
            N=4   & 26.9  & 59.2  & 8.9   & 21.7 \\
            N=6   & 28.2  & 61.9  & 9.8   & 22.6 \\
            N=8 (ours) & \textbf{29.4} & \textbf{63.2} & \textbf{10.4} & \textbf{23.7} \\
            N=10  & 28.4  & 63.1  & 10.4  & 22.8 \\
            N=12  & 27.7  & 60.8  & 10.0  & 22.5 \\
            \bottomrule
            \end{tabular}%
    }
    \hspace{3mm}
	\subfloat[Study on the input gait position $p$ in GSP. \label{tab:GSP_position}]{
		\tablestyle{3pt}{0.88}
		    \begin{tabular}{ccccc}
            \toprule            \multirow{3}[6]{*}{Methods} & \multicolumn{4}{c}{LTCC} \\
        \cmidrule{2-5}          & \multicolumn{2}{c}{Standard} & \multicolumn{2}{c}{Cloth-Changing} \\
        \cmidrule{2-5}          & mAP   & Rank-1 & mAP   & Rank-1 \\
            \midrule
            Baseline & 23.2  & 55.1  & 8.1   & 19.6 \\
            Arb.  & 27.1  & 59.5  & 9.2   & 20.5 \\
            BEGN  & 28.4  & 61.2  & 9.8   & 22.0 \\
            END   & 28.1  & 61.5  & 9.5   & 22.4 \\
            Mid. (ours) & \textbf{29.4} & \textbf{63.2} & \textbf{10.4} & \textbf{23.7} \\
            \bottomrule
            \end{tabular}%
                }
                \hspace{3mm}
	\subfloat[Study on the used losses in SC constraint.\label{tab:SC_Loss}]{
		\tablestyle{3pt}{0.99}
        \begin{tabular}{ccccc}
            \toprule            \multirow{3}[6]{*}{Methods} & \multicolumn{4}{c}{LTCC} \\
        \cmidrule{2-5}          & \multicolumn{2}{c}{Standard} & \multicolumn{2}{c}{Cloth-Changing} \\
        \cmidrule{2-5}          & mAP   & Rank-1 & mAP   & Rank-1 \\
            \midrule
            Baseline & 23.2  & 55.1  & 8.1   & 19.6 \\
            w/ $\mathcal{L}_{MSE}$ & 27.5  & 61.0  & 9.0   & 21.4 \\
            w/o $\mathcal{L}_{recon.}$ & 28.3  & 62.7  & 9.6   & 22.9 \\
            ours  & \textbf{29.4} & \textbf{63.2} & \textbf{10.4} & \textbf{23.7} \\
            \bottomrule
            \end{tabular}%
	}
	\vspace{-5mm}
	\label{tab:choices}
\end{table*}

\begin{table}[htbp]
  \centering
  \footnotesize
  \caption{Study on the different ReID inference strategies.}
  \vspace{-2mm}
  \setlength{\tabcolsep}{4mm}{
  \tablestyle{12.6pt}{0.8}
    \begin{tabular}{ccccc}
    \toprule
    \multirow{3}[6]{*}{Methods} & \multicolumn{4}{c}{LTCC} \\
\cmidrule{2-5}          & \multicolumn{2}{c}{Standard} & \multicolumn{2}{c}{Cloth-Changing} \\
\cmidrule{2-5}          & mAP   & Rank-1 & mAP   & Rank-1 \\
    \midrule
    Baseline & 23.2  & 55.1  & 8.1   & 19.6 \\
    $\widetilde{R}$ & 8.6   & 21.1  & 4.3   & 9.9 \\
    $\hat{r} + \hat{g}$ & \textbf{29.8} & \textbf{64.0} & \textbf{10.9} & \textbf{24.4} \\
    $\widetilde{r} + \widetilde{g}$ & 28.9  & \underline{63.2}  & 9.7   & 23.1 \\
    $\widetilde{r}$ & 28.1  & 60.8  & 9.1   & 21.3 \\
    $r$ (ours) & \underline{29.4}  & \underline{63.2}  & \underline{10.4}  & \underline{23.7} \\
    \bottomrule
    \end{tabular}}%
    \vspace{-5.5mm}
  \label{tab:infer}%
\end{table}%


\subsection{Design Choices in Our GI-ReID Framework}

We study the different design choices in our GI-ReID framework. We train and test model on the real large-scale cloth-changing ReID dataset LTCC~\cite{qian2020long}.



\noindent\textbf{Influence of the Length $N$ of Predicted Gait Sequence.} As shown in Eq-(\ref{eq:loss_pred}) of Sec.~\ref{sec:GSP}, the output of GSP $\widetilde{R} \in \mathbb{R}^{N*h*w}$ is a sequence with $N$ predicted gait frames. We study the influence of length $N$ w.r.t the ReID performance. Table~\ref{tab:GSP_len} shows that when $N=8$, our GI-ReID gets the best performance, achieving a good trade-off between gait prediction error and gait information augmentation.

\noindent\textbf{Is `Middle Frame Input Principle' Necessary?} As described in Eq-(\ref{eq:loss_position}) of GSP in Sec.~\ref{sec:GSP}, we employ a position embedder $P$ and a feature aggregator $A$ to set up a \emph{middle frame input principle} to reduce the gait prediction ambiguity and difficulty. Here we compare several schemes to show the necessity of such design. \emph{\textbf{Arb.}}: we remove position embedder $P$, feature aggregator $A$, position loss $\mathcal{L}_{position}$ for GSP, and take the gait silhouette at \emph{arbitrary} position as input for training. \emph{\textbf{BEGN}} and \emph{\textbf{END}}: we respectively take the gait stance at the \emph{beginning} and the \emph{end} position as input to predict gait sequence during the GSP training. In Table~\ref{tab:GSP_position}, the scheme~\emph{\textbf{Mid. (ours)}} that uses the gait frame at middle position for gait sequence prediction achieves the best performance, outperforming \emph{Arb.} by 2.3\% in mAP in the standard setting, which reveals that predicting the gait statuses \textbf{before and after} the input middle gait status indeed could reduce prediction difficulty/ambiguity. 


\noindent\textbf{Why Use MMD for Regularization?} For the SC constraint, we shrink the gap between the embeded ReID vector $\hat{r}$ and gait vector $\hat{g}$ by minimizing MMD through $\mathcal{L}_{MMD}$. We study this design in Table~\ref{tab:SC_Loss} and find that when replacing $\mathcal{L}_{MMD}$ with $\mathcal{L}_{MSE}$, the performance of \emph{w/ $\mathcal{L}_{MSE}$} drops nearly 2.0\% in mAP. That's because MMD loss is a distribution-level constraint and could better enforce the high-level semantics consistency between dynamic motion gait features and static spatial ReID features. MSE loss is an element-wise constraint, and not so suitable to coordinate two modalities of motion gait and RGB feature.

\noindent\textbf{Is Reconstruction Penalty Necessary?} When removing $\mathcal{L}_{recon.}$ in Eq-(\ref{eq:loss_recon}), as shown in Table~\ref{tab:SC_Loss}, the shceme \emph{w/o $\mathcal{L}_{recon.}$} is inferior to ours by 1.1\%/0.8\% in mAP in the two settings, which demonstrates that avoiding information lost caused by feature regularization could enhance the final ReID performance of our GI-ReID framework.




\noindent\textbf{Which One for ReID Inference?} We compare several cases of using (1) predicted gait sequence $\widetilde{R}$, (2) aligned features fusion $\hat{r}$ + $\hat{g}$, (3) reconstructed features fusion $\widetilde{r}$ + $\widetilde{g}$, and (4) reconstructed ReID vector $\widetilde{r}$ for ReID inference. Table~\ref{tab:infer} shows that 1) Directly using the predicted gait sequence $\widetilde{R}$ for CC-ReID failed to get satisfactory results, this also indicates that these gait recognition works~\cite{chao2019gaitset,Fan_2020_CVPR,elharrouss2020gait} are not optimal for CC-ReID. 2) Using the well-aligned features fusion $\hat{r} + \hat{g}$ achieves the best performance, outperforming ours by 0.4\%/0.5\% in mAP in the two settings, but this scheme still needs Gait-Stream in the inference. 3) Using the reconstructed ReID vector $\widetilde{r}$ for inference suffers from information lost and is inferior to ours by 1.3\% in mAP in the both two settings. 4) Our scheme that using the regularized ReID vector $r$ achieves the second best performance while saving the computation costs brought by Gait-Stream.


\begin{figure}[h]
  \vspace{-3.mm}
  \centerline{\includegraphics[width=1.0\linewidth]{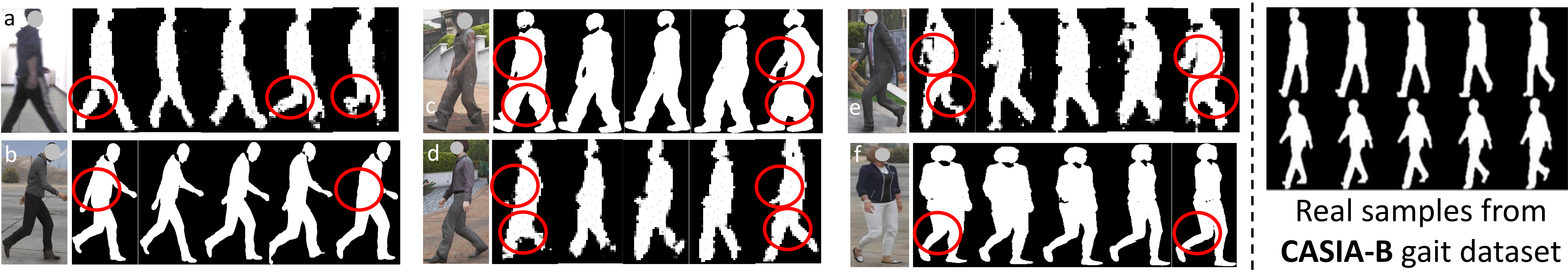}}
  \vspace{-3mm}
  \caption{Six predicted gait sequences \textit{vs.} realistic gait samples.}
  \label{fig:vis_gait}
  \vspace{-4.5mm}
\end{figure}

\subsection{More Analysis, Visualization and Insights}

To further prove that the proposed gait sequence prediction (GSP) module can actually predict unique human motion features, and the achieved improvements of GI-ReID indeed come from gait information, not from using additional gait-related datasets or person silhouette images, here we provide more analysis and visualization results. For example, when testing the gait-based recognition performance using GaitSet~\cite{chao2019gaitset} on the predicted human gait sequences $\widetilde{R}$ generated by GSP module (see \textbf{Supplementary} for details), it can achieve a competitive 62.4\% in Rank-1 on CASIA-B.

\begin{table*}[t]\centering
\vspace{-3mm} 
    \caption{Performance (\%) comparisons of our GI-ReID and other competitors on the cloth-changing datasets LTCC~\cite{qian2020long} and PRCC~\cite{yang2019person}. `$\dag$' means that only identities with clothes changing are used for training. More results are presented in \textbf{Supplementary}.}
    \vspace{-3mm}
	\captionsetup[subffloat]{justification=centering}
	\subfloat[Comparison results on LTCC.\label{tab:LTCC}]{
	\tablestyle{1.6pt}{1.11}
        \begin{tabular}{@{\extracolsep{\fill}}l|c|c|c|c||c|c|c|c}
        \hline 
        \multicolumn{1}{c|}{\multirow{2}{*}{Methods}} & \multicolumn{2}{c|}{Standard} & \multicolumn{2}{c||}{Cloth-changing} & \multicolumn{2}{c|}{Standard$^\dag$} & \multicolumn{2}{c}{Cloth-changing$^\dag$}\tabularnewline
        \cline{2-9} 
         & Rank-1 & mAP & Rank-1 & mAP & Rank-1 & mAP & Rank-1 & mAP\tabularnewline
        \hline 
        LOMO \cite{XQDA} + NullSpace \cite{NullReid} & 34.83 & 11.92 & 16.45 & 6.29 & 27.59 & 9.43 & 13.37 & 5.34 \tabularnewline
        
        ResNet-50 + Face~\cite{xue2018clothing} & 60.44 & 25.42 & 22.10 & 9.44 & 55.37 & 22.23 & 20.68 & 8.99
        \tabularnewline
        PCB \cite{sun2018beyond} & 65.11 & 30.60 & 23.52 & 10.03 & 59.22 & 26.61 & 21.93 & 8.81 \tabularnewline
        HACNN \cite{li2018harmonious} & 60.24 & 26.71 & 21.59 & 9.25 & 57.12 & 23.48 & 20.81 & 8.27 \tabularnewline
        MuDeep \cite{qian2019leader} & 61.86 & 27.52 & 23.53 & 10.23 & 56.99 & 24.10 & 18.66 & 8.76\tabularnewline
        
        
        \hline
        Baseline (ResNet-50) & 55.14 & 23.21 & 19.58 & 8.10 & 54.27 & 21.98 & 19.14 & 7.74 \\
        
        GI-ReID (ResNet-50, ours) & 63.21 & 29.44 & 23.72 & 10.38 & 61.39 & 27.88 & 22.59 & 9.87  \\
        \hline
        Baseline (OSNet) & 66.07 & 31.18 & 23.43 & 10.56  & 61.22 & 27.41 & 22.97 & 9.74 \\
        
        GI-ReID (OSNet, ours) & \bf 73.59 & \bf36.07 & {28.11} & {13.17} & \bf 66.94 & \bf 33.04 & \bf 26.71 & {12.69} \\
        \hline 
        
        Baseline (LTCC-shape~\cite{qian2020long}) & -- & -- & {26.15} & {12.40} & -- & -- & {25.15} &  {11.67} \\
        
        LTCC-shape + Gait-Stream (ours) & -- & -- & \bf 28.86 & \bf 14.19 & -- & -- & {26.41} &  \bf 13.26 \\
        \hline
        \end{tabular}%
    }
    \hspace{0.5mm}
	\subfloat[Comparison results on PRCC.\label{tab:PRCC}]{
		\tablestyle{2.0pt}{0.9}
		    \begin{tabular}{c|c|c|c}
            \hline
            \multirow{2}[1]{*}{Methods} & \multicolumn{3}{c}{Cross-clothes} \\
        \cline{2-4}          & Rank-1 & Rank-10 & Rank-20 \\
            \hline
            Shape~\cite{belongie2002shape} & 11.48 & 38.66 & 53.21 \\
            LNSCT~\cite{xie2010extraction} & 15.33 & 53.87 & 67.12 \\
            HACNN~\cite{li2018harmonious} & 21.81 & 59.47 & 67.45 \\
            PCB~\cite{sun2018beyond} & 22.86 & 61.24 & 78.27 \\
            SketchNet~\cite{zhang2016sketchnet} & 17.89 & 43.70  & 58.62 \\
            Deformable~\cite{dai2017deformable} & 25.98 & 71.67 & 85.31 \\
            STN~\cite{jaderberg2015spatial} & 27.47 & 69.53 & 83.22 \\
            
            RCSANet~\cite{huang2021clothing} & 31.60 & -- & -- \\
            
            \hline
            PRCC-contour~\cite{yang2019person} & {34.38} & {77.30}  & {88.05} \\

            + Gait-Stream (ours)  & {36.19} & {79.93}   & {91,67}    \\
            \hline
            Baseline (ResNet-50) & 22.23 & 61.08 & 76.44 \\
            GI-ReID (ResNet-50) & 33.26 & 75.09 & 87.44 \\
            \hline
            Baseline (OSNet) & 28.70 & 72.34 & 85.89 \\
            GI-ReID (OSNet) & \textbf{37.55} & \textbf{82.25} & \textbf{93.76} \\
            \hline
            \end{tabular}%
                }
	\vspace{-5mm}
	\label{tab:LTCC_PRCC}
\end{table*}

\noindent\textbf{Gait Sequence Prediction Visualization.} Figure~\ref{fig:vis_gait} further shows 6 groups of gait prediction results (left) and 2 groups of realistic gait samples from CASIA-B dataset~\cite{chao2019gaitset} (right). Compared to the realistic gait samples, the predicted gait results (\ieno, the outputs of GSP) have the reasonable continuous movements, \egno, swing arms and opening/closing legs. The gait-stream could learn the discriminative dynamic clues from these predicted gait results, like \textit{the walking stride}, \textit{the left-right swinging range of arms}, \textit{the opening/closing angle of legs}, \etcno, (see red circles in Figure~\ref{fig:vis_gait}).



\begin{figure}
  \centerline{\includegraphics[width=1.0\linewidth]{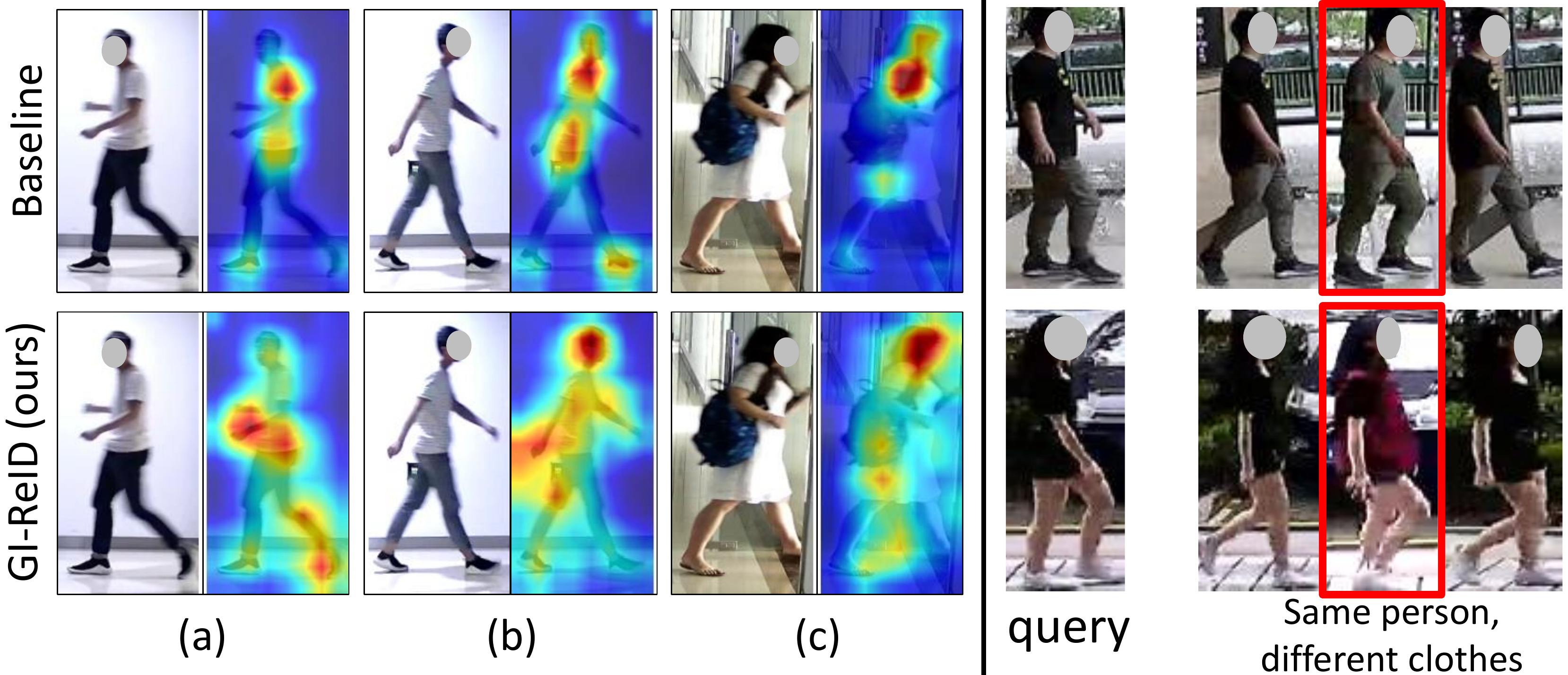}}
  \vspace{-3.5mm}
  \caption{\textbf{Left:} three examples of activation maps comparison between baseline and our GI-ReID, which shows GI-ReID not only focuses on people's clothes, but also pay attention to the holistic human gait and local face; \textbf{Right:} Top-3 ranking list of GI-ReID for two query images on Real28. GI-ReID could identify the same person with different clothes based on the assistance of gait.}
  \vspace{-6mm}
  \label{fig:vis_feat}
\end{figure}

\noindent\textbf{Feature Map Visualization.} To better understand how our GI-ReID works, we visualize the intermediate activation feature maps of \emph{Baseline} and our GI-ReID for comparison following~\cite{zhou2019omni,jin2020global,jin2020style}. On the left of Figure~\ref{fig:vis_feat}, we show three examples of activation maps on Real28, and we observe that the feature maps of \emph{Baseline} have high response mainly on person's clothes. In contrast, the activation features of our GI-ReID not only have high response on person's clothes, but also cover the holistic human body structure (gait) and local face information (robust to cloth changing).


\vspace{-1mm}
\subsection{Comparison with State-of-the-Arts}
The study on cloth-changing ReID is relatively rare~\cite{huang2019beyond,huang2019celebrities,yang2019person,qian2020long,li2020learning,yu2020cocas,wan2020person}, and most of them have not released source codes, even the dataset~\cite{yu2020cocas}. We compare our GI-ReID with multiple general ReID algorithms, including PCB~\cite{sun2018beyond}, HACNN~\cite{li2018harmonious}, MuDeep~\cite{qian2019leader}, and specific cloth-changing ReID methods LTCC-shape~\cite{qian2020long}, PRCC-contour~\cite{yang2019person}, RCSANet~\cite{huang2021clothing}. In Table~\ref{tab:LTCC_PRCC}, we observe that 1) Thanks to the cloth-independent gait characteristics, our scheme GI-ReID (OSNet) achieves the best performance on PRCC, outperforming the second best PRCC-contour~\cite{yang2019person} by 3.17\% in Rank-1 in the cross-clothes setting. 2) The proposed Gait-Stream, as a kind of regularization, could benefit other methods, \egno, LTCC-shape~\cite{qian2020long}. We find that the scheme of \emph{LTCC-shape + Gait-Stream} could further obtain 1.79\%/1.59\% gain in mAP on LTCC. 3) For two cloth-changing settings of LTCC, our scheme GI-ReID (ResNet-50) both achieve obvious gains (2.28\%/2.13\% in mAP) over the Baseline (ResNet-50), which totally-fair results indicate that GI-ReID could handle clothes changes and learn identity-relevant features. 4) Our method is compatible with existing ReID networks, \egno, built upon the strong ReID-specific network OSNet~\cite{zhou2019omni}, GI-ReID (OSNet) further achieves gains than GI-ReID (ResNet-50). 


\subsection{Failure Cases Analysis}
Due to the large difference on the capture viewpoints and environments between gait and ReID training data, the predicted gait results of GSP are not always perfect (see \textbf{Supplementary}) when occlusion, partial, multi-person, \etcno, existed in the person images, which may affect the CC-ReID performance. That is why we indirectly use gait predictions in a knowledge regularization manner, which makes GI-ReID robust and not sensitive to these failure cases.





\vspace{-1mm}
\section{Conclusion}
In this paper, we propose to utilize human unique gait to address the cloth-changing ReID problem from a single image. A novel gait-involved two-stream framework GI-ReID is introduced, which takes gait as a regulator with a Gait-Stream (discarded in the inference), to encourage the cloth-agnostic representation learning of image-based ReID-Stream. To facilitate the gait utilization, a gait sequence prediction (GSP) module and a high-level semantics consistency (SC) constraint are further designed. Extensive experiments on multiple CC-ReID benchmarks demonstrate the effectiveness and superiority of GI-ReID.

\section{Acknowledgements}

This work was (partially) supported by the National Key R\&D Program of China under Grant 2020AAA0103902, Alibaba Innovative Research (AIR) program, and NSFC under Grant U1908209, 61632001, 62021001.


\appendix
  \renewcommand\thesection{\arabic{section}}

\vspace{4mm}
\noindent{\LARGE \textbf{Supplementary}}


\section{Detailed Network Structures of GI-ReID}

GI-ReID, as a image-based cloth-changing ReID framework, with gait information as assistance, consists of an auxiliary Gait-Stream and a mainstream ReID-Stream. ReID-Stream can be arbitrary commonly-used network architectures, such as ResNet~\cite{he2016deep}, and also can be some ReID-specific network architectures, such as PCB~\cite{sun2018beyond}, OSNet~\cite{zhou2019omni}. Thus, in this section, we mainly introduce/describe the detailed network architecture of Gait-Stream, which contains two key parts, GSP module for gait information prediction/augmentation and GaitSet~\cite{chao2019gaitset} for gait features extraction.

\textbf{Architecture of Gait Sequence Prediction (GSP) Module:} The proposed GSP module consists of a feature encoder $E$, a decoder $D$, a position embedder $P$, and a feature aggregator $A$.

\noindent\textbf{(1). Encoder $E$.} The encoder $E$ is a CNN with four Conv. layers (filter size = $4 \times 4$ and stride = 2). The number of filters is increased from 64 $\rightarrow$ 512. Each Conv. layer is followed by a batch-normalization (BN) layer~\cite{ioffe2015batch} and a rectified linear unit (ReLU) activation function~\cite{nair2010rectified}. In the end, a 100-dimensional feature is obtained through a fully connected (FC) layer.

\begin{figure}
  \centerline{\includegraphics[width=0.9\linewidth]{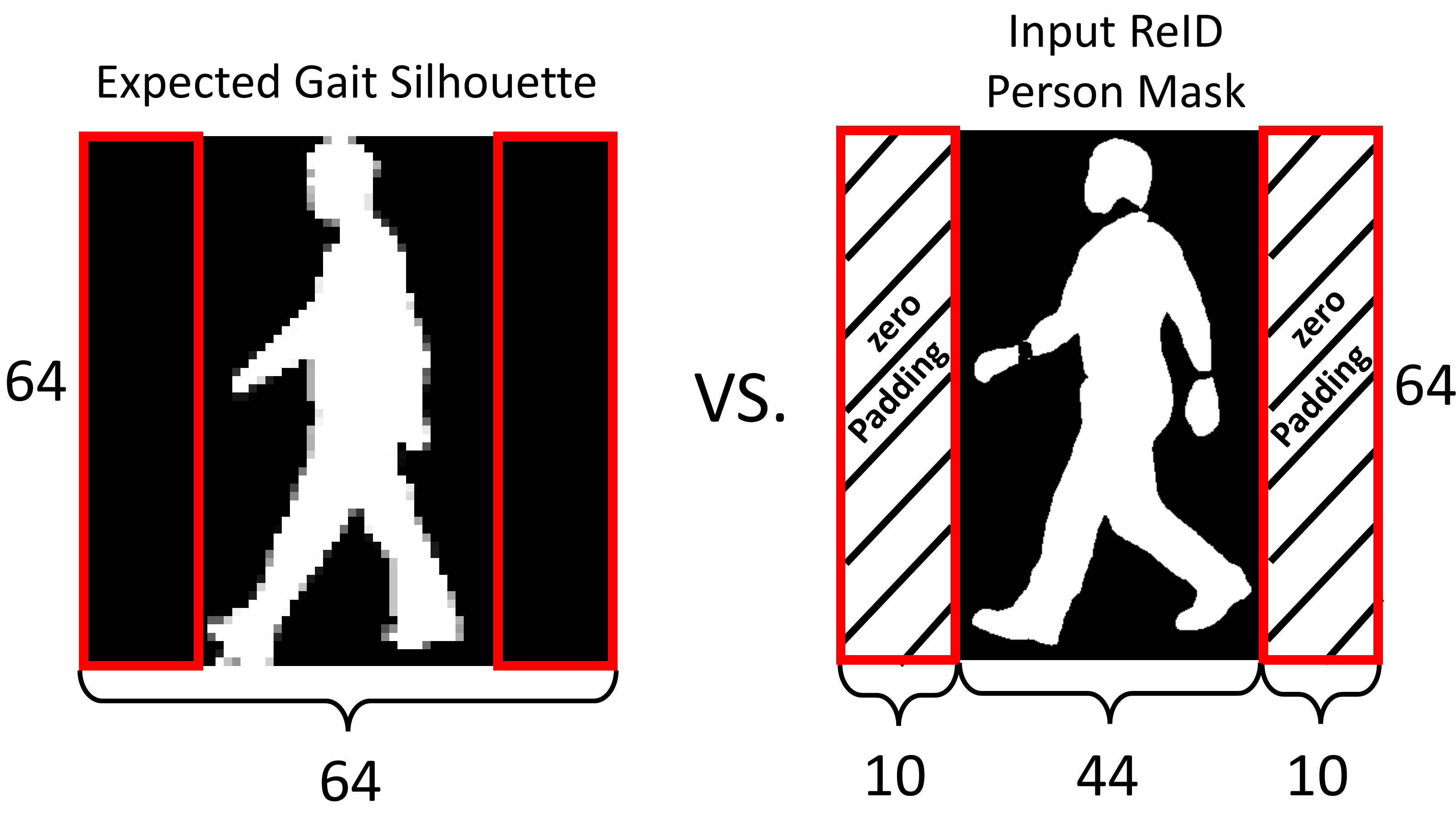}}
  \vspace{-3.5mm}
  \caption{We apply ``resize+zero\_padding'' in the person masks (right) when fine-tuning GSP module on the ReID-specific datasets, because the raw gait training data (left) typically have the height-width ratio of (1:1), which is important/necessary for training GSP to get satisfactory gait prediction results.}
  \vspace{-3mm}
  \label{fig:supp_gsp_input}
\end{figure}

Note that, when pre-training GSP module on the gait-specific datasets following~\cite{chao2019gaitset}, the input gait silhouette of encoder $E$ has a size of $1 \times 64 \times 64$ (height-width ratio is 1:1). We use CASIA-B~\cite{chao2019gaitset} as training dataset. On the ReID-specific datasets, since the input person images usually have a height-width ratio of 2:1 (\egno, height-256, width-128), we need leverage an operation of ``resize+zero\_padding'' to handle such training data gap, which is pivotal for GSP's accurate gait sequence prediction. For better understanding, we vividly visualize such process in Figure~\ref{fig:supp_gsp_input}.

\noindent\textbf{(2). Position Embedder $P$ and Feature Aggregator $A$.} To reduce the gait prediction ambiguity and difficulty of GSP, a position embedder $P$ and a feature aggregator $A$ are introduced to integrate the prior information of input middle frame index into the prediction process of GSP. The position embedder $P$ has a similar structure to that of the encoder $E$, but with one more FC layer to regress the 1D position label $\widetilde{p}$. The feature aggregator $A$ is inserted between the encoder and the decoder to convert the raw encoded features $f_S$ into middle-position-aware features $f_{S}^{\widetilde{p}}$ by taking the embedded middle position information $\widetilde{p}$ into account. With respect to the architecture of $A$, it is implemented only by a FC layer, which aims to regress to the aggregated 100-dimension feature $f_{S}^{\widetilde{p}} \in \mathbb{R}^{100}$ from the \textbf{101-dimension} concatenated vector of the raw encoded feature $f_S \in \mathbb{R}^{100}$ and the embedded middle position prior vector $\widetilde{p} \in \mathbb{R}^1$.

\noindent\textbf{(3). Decoder $D$.} The structure of the decoder $D$ is symmetrical to that of the encoder $E$. A FC layer along with reshaping is first employed to convert the input 100D feature into the same size as the last feature output of the encoder $E$, and then four DeConv. layers are used for up-sampling. A sigmoid activation function is applied in the end, and
outputs the gait predictions with a size of $N \times 64 \times 64$, where each channel
indicates a predicted gait frame of final results.

\textbf{Architecture of GaitSet:} GaitSet~\cite{chao2019gaitset} is a classic set-based gait recognition network, which takes a set of silhouettes/gait frames as input. After obtaining features from each input silhouette independently using a CNN, \emph{set pooling} is applied to merge features over frames into a set-level feature. This set-level feature is then used for discrimination learning via
\emph{horizontal pyramid mapping} (HPM), which aims to extract features of different spatial locations on different scales. We recommend seeing more details from their original paper~\cite{chao2019gaitset}.

\section{Training Details of our GI-ReID}

\noindent\textbf{Phase-1: Pre-training for GaitSet.} The input is a set of aligned silhouettes in size of $64 \times 44$. The silhouettes are directly provided by the datasets and are aligned based on methods in \cite{takemura2018multi}. The set cardinality in the training is set to be 30. Adam is chosen as an optimizer. The number of scales $S$ in HPM is set as $5$. The margin in separate triplet loss $\mathcal{L}_{tri}^{sep}$~\cite{chao2019gaitset} is set as $0.2$. The mini-batch is composed of $P=16$ and $N=8$ ($P, N$ respectively mean the number of person identities and input gait frames). We set the number of channels in $C1$ and $C2$ as $32$, in $C3$ and $C4$ as $64$ and in $C5$ and $C6$ as 128 (following~\cite{chao2019gaitset}). The learning rate is set to be $1\times$10$^{-4}$, and the model is trained for 80 epochs.
 
\noindent\textbf{Phase-2: Joint Training for GSP module and GaitSet.} After pre-training GaitSet, we jointly train the proposed gait sequence prediction (GSP) module and GaitSet for Gait-Stream. Specifically, during the joint-training, we also re-use CASIA-B dataset for effective gait prediction training. Following~\cite{hermans2017defense}, a batch is formed by first randomly sampling $P$ identities. For each identity, we sample $N$ continuous gait frames as the ground-truth gait sequence. Then the batch size is $B=P\times N$. We set $P=4$ and $N=8$ (\ieno, batch size $B=P\times N=32$. As presented in the main manuscript, we use the middle one of the ground-truth gait sequence (\ieno, the fourth one when $N=8$) as input for GSP training. We first optimize GSP with the proposed position loss $\mathcal{L}_{position}$ and prediction loss $\mathcal{L}_{pred}$ (loss balance is set as 1:1) for 80 epochs, which enables GSP to output reasonable predicted gait sequence results. We train GSP with Adam optimizer \cite{kingma2014adam} with a initial learning rate of 5$\times$10$^{-4}$. We optimize the Adam optimizer with a weight decay of 1$\times$10$^{-4}$. The learning rate is decayed by a factor of 0.1 at 40 epoch.

\begin{algorithm}[htp]
   \caption{Training Process of GI-ReID}
   \label{alg:GI-ReID}
   \footnotesize
\begin{algorithmic}[1] 
   \STATE {\bf Input:} gait dataset $\mathcal{G}$ (\egno, CASIA-B~\cite{chao2019gaitset}), ReID dataset $\mathcal{R}$ (\egno, LTCC~\cite{qian2020long}). Learning rate is simply denoted as $\eta$. The entire GI-ReID framework consists of GSP module $GSP_{\theta}$, GaitSet (GS) $GS_{\phi}$, SC constraints related FC layers $SC_{\psi}$, and ReID-Stream backbone $ReID_{\omega}$. 
   \STATE {\bf Output:} inference ReID vector $r$.
   
    \STATE \textcolor{red}{\#\#\# Phase-1: Pre-training for GaitSet}
    \FOR{$epoch = 1$ {\bf to} $80$}
    \STATE Sample $P\times N = 16\times 8$ samples from gait training set $\mathcal{G}$.
    \STATE $\mathcal{L}_{total} = \mathcal{L}_{tri}^{sep}$ \hfill // \textit{Use the separate triplet loss as objective function~\cite{chao2019gaitset}.}
    \STATE $\phi = \phi - \eta \nabla_{\phi} \mathcal{L}_{tri}^{sep}$ \hfill // \textit{Update GaitSet (GS) $GS_{\phi}$.}  
   \ENDFOR

    \STATE \textcolor{red}{\#\#\# Phase-2: Joint Training for GSP module and GaitSet}
    
    \FOR{$epoch = 1$ {\bf to} $80$}
    \STATE Sample $P\times N = 4\times 8$ samples from gait training set $\mathcal{G}$.
    \STATE $\mathcal{L}_{total} = \mathcal{L}_{position} + \mathcal{L}_{pred}$ \hfill // \textit{Use the proposed position loss and prediction loss as objective functions.}
    \STATE $\theta = \theta - \eta \nabla_{\theta} \mathcal{L}_{total}$ \hfill // \textit{Warm up GSP module $GSP_{\theta}$.}  
   \ENDFOR
   
    \FOR{$epoch = 1$ {\bf to} $160$}
    \STATE Sample $P\times N = 4\times 8$ samples from gait training set $\mathcal{G}$.
    \STATE $\mathcal{L}_{total} = \mathcal{L}_{position} + \mathcal{L}_{pred} + \mathcal{L}_{tri}^{sep}$ \hfill // \textit{Use the position loss, prediction loss , and separate triplet loss as objective functions.}
    \STATE $(\theta, \phi) = (\theta, \phi) - \eta \nabla_{(\theta, \phi)} \mathcal{L}_{total}$ \hfill // \textit{Jointly update GSP module $GSP_{\theta}$ and GaitSet (GS) $GS_{\phi}$.}  
   \ENDFOR
   
    \STATE \textcolor{red}{\#\#\# Phase-3: Joint Training for Gait-Stream and ReID-Stream}
    
    \FOR{$epoch = 1$ {\bf to} $240$}
    \STATE Sample $P\times N = 10\times 8$ samples from ReID training set $\mathcal{R}$.
    \STATE $\mathcal{L}_{total} = 0.1*\mathcal{L}_{position} + 0.1*\mathcal{L}_{pred} +
    0.1*\mathcal{L}_{tri}^{sep} +
    \mathcal{L}_{cla} + 
    \mathcal{L}_{tri}^{HM} +
    0.5*\mathcal{L}_{MMD} + 
    0.5*\mathcal{L}_{recon.}$\hfill // \textit{Total objective functions consists of the position loss, prediction loss, separate triplet loss (for Gait-Stream), and the classification loss, triplet loss (with hard-mining, HM) (for ReID-Stream), and the MMD loss, reconstruction loss (SC constraints).}
    \STATE $(\theta, \phi, \psi, \omega) = (\theta, \phi, \psi, \omega) - \eta \nabla_{(\theta, \phi, \psi, \omega)} \mathcal{L}_{total}$ \hfill // \textit{Jointly update GSP module $GSP_{\theta}$, GaitSet (GS) $GS_{\phi}$, SC constraints related FC embedding layers $SC_{\psi}$, and ReID-Stream backbone $ReID_{\omega}$.}
   \ENDFOR
\end{algorithmic}
\end{algorithm}

After warming up the GSP module for 80 epochs, we jointly train GSP and GaitSet for extra 160 epochs with initial learning rate as 5$\times$10$^{-4}$. We also use Adam optimizer \cite{kingma2014adam} for optimization with a weight decay of 1$\times$10$^{-4}$, the learning rate is decayed by a factor of 0.5 at 40, 80, and 120 epochs. When jointly training GSP and GaitSet, excluding the GSP-related position loss $\mathcal{L}_{position}$ and prediction loss $\mathcal{L}_{pred}$, we further use \emph{separate triplet loss} $\mathcal{L}_{tri}^{sep}$ for training, which is introduced in details in GaitSet~\cite{chao2019gaitset}, and we also set the loss weight as 1.0 for this supervision.

\noindent\textbf{Phase-3: Joint Training for Gait-Stream and ReID-Stream.} When we jointly training Gait-Stream and
ReID-Stream on the ReID datasets, Gait-Stream is also fine-tuned/learnable. Since the full gait sequence ground truth are not available for ReID-specific datasets, we adjust the original prediction loss $\mathcal{L}_{pred}$ in GSP by only calculating L1 distance between the single input person mask and the middle frame result
of the entire predicted gait sequence.

On the large-scale cloth-changing datasets VC-Clothes~\cite{wan2020person}, LTCC~\cite{qian2020long}, and PRCC~\cite{yang2019person}, we set training batch size as $B=80=P\times N=10\times 8$. Both of Gait-Stream (including GSP and GaitSet) and ReID-Stream use Adam optimizer \cite{kingma2014adam} for optimization, where the initial learning rate for Gait-Stream is 1$\times$10$^{-5}$, for ReID-Stream is 5$\times$10$^{-4}$. We optimize two Adam optimizers for Gait-Stream and ReID-Stream with a weight decay of 1$\times$10$^{-5}$ for total 240 epochs. The learning rate is decayed by a factor of 0.1 at 80 and 160 epochs for ReID-Stream, while no learning rate decay for Gait-Stream. For the losses usage, we adopt the widely-adopted classification loss $\mathcal{L}_{cla}$~\cite{sun2018beyond,fu2019horizontal}, and triplet loss with batch hard mining $\mathcal{L}_{tri}^{HM}$\cite{hermans2017defense}) as basic optimization objectives for ReID-Stream training, and we set these two loss weights as 1.0. Besides, for the Gait-Stream related losses, including $\mathcal{L}_{position}, \mathcal{L}_{pred}, \mathcal{L}_{tri}^{sep}$, we set all their loss weights as 0.1. For the semantics consistency (SC) constraints related FC embedding layers, we merge their learnable parameters into ReID-Stream's optimization, and set the balance weights for MMD loss $\mathcal{L}_{MMD}$ and reconstruction penalty $\mathcal{L}_{recon.}$ as 0.5. The pseudo code of the entire training process of our GI-ReID is given in Algorithm~\ref{alg:GI-ReID}.

\section{Details of Datasets}

We use one widely-used video ReID dataset MARS~\cite{zheng2016mars}, and four image-based cloth-changing ReID datasets Real28~\cite{wan2020person}, VC-Clothes~\cite{wan2020person}, LTCC~\cite{qian2020long}, PRCC~\cite{yang2019person} to perform experiments. In Table~\ref{tab:datasets}, we present the detailed information about these ReID datasets. 

\begin{table}[htb]
  \footnotesize
  \vspace{-1mm}
  \centering
  \caption{Brief introduction/comparison of datasets.}
  \vspace{-2mm}
  \setlength{\tabcolsep}{0.6mm}{ 
    \begin{tabular}{c|c|c|c|c|c}
    \toprule
          & MARS  & Real28 & VC-Clothes & LTCC & PRCC \\
    \hline
    Category & Video & Image & Image & Image & Image\\
    Photo Style & Real  & Real  & Synthetic & Real & Real \\
    Scale & Large  & Small  & Large & Large & Large \\    
    Cloth Change & No    & Yes   & Yes   & Yes & Yes \\
    Identities & 1,261 & 28    & 512   & 152 & 221 \\
    Samples & 20,715 & 4,324 &  19,060 & 17,138 & 33,698\\
    Cameras & 6     & 4     & 4     & N/A & 3\\
    Usage & Train\&Test & Test  & Train\&Test & Train\&Test & Train\&Test\\
    \bottomrule
    \end{tabular}}%
    \vspace{-2mm}
  \label{tab:datasets}%
\end{table}%

\emph{MARS} is a popular dataset for video-based person ReID. There are 20,715 track-lets come from 1,261 pedestrians who are captured by at least 2 cameras. We use the train/test split protocol defined in~\cite{zheng2016mars}. 

\emph{Real28, VC-Clothes, LTCC and PRCC} are all newly released image datasets for cloth-changing ReID~\cite{wan2020person,qian2020long,yang2019person}. 

\emph{Real28} is a \textbf{small} real-scenario dataset, which is collected in 3 different days (with different clothing) by 4 cameras. It consists of totally 4,324 images from 28 different identities with 2 indoor scenes and 2 outdoors. Similar to~\cite{wan2020person}, since the size of Real28 is not big enough for training deep learning models, we just use it for evaluation. There are totally 336 images in the query and 3,988 images in the gallery.

\emph{VC-Clothes} is a \textbf{synthetic} dataset where images are rendered by the Grand Theft Auto V (GTA5). It has 512 identities, 4 scenes (cameras) and on average 9 images/scenes for each identity and a total number of 19,060 images. Following~\cite{wan2020person}, we equally split the dataset by identities: 256 identities for training and the other 256 for testing. We randomly chose 4 images per person from each camera as query, and have the other images serve as gallery images. Eventually, we get totally 9,449 images in the training, 1,020 images as queries and 8,591 others in the gallery.

\emph{LTCC} is a \textbf{large-scale real-scenario} cloth-changing dataset, which contains 17,138 person images of 152 identities. On average, there are 5 different clothes for each
cloth-changing person, with the numbers of outfit changes ranging from 2 to 14. Following~\cite{qian2020long}, we split the LTCC dataset into training and testing sets. The training set consists of 77 identities, where 46 people have cloth changes and the rest of 31 people wear the same outfits during the recording. Similarly, the testing set contains 45 people with changing clothes and 30 people wearing the same outfits.


\emph{PRCC} is also a \textbf{large-scale real-scenario} cloth-changing dataset, recently published by Yang~\etal~\cite{yang2019person}. It consists of 221 identities with three camera views \emph{Camera A}, \emph{Camera B}, and \emph{Camera C}. Each person in Cameras A and B is wearing the same clothes, but the images are captured in different rooms. For Camera C, the person wears different clothes, and the images are captured in a different day. The images in the PRCC dataset include not only clothing changes for the same person across different camera views but also other variations, \egno~ changes in illumination, occlusion, pose and viewpoint. In summary, nearly 50 images exists for each person in each camera view. Therefore, approximately 152 images of each person are included in the dataset, for 33,698 images in total. 

Following~\cite{yang2019person}, we split the PRCC dataset into a training set and a testing set. The training set consists of 150 people, and the testing set consists of 71 people, with no overlap between them in terms of identities. The testing set was further divided into a gallery set and a probe set. For each identity in the testing set, we chose one image in \emph{Camera view A} to form the gallery set for a single-shot matching. All images in \emph{Camera views B} and \emph{Camera C} were used for the probe set. Specifically, the person matching between Camera views A and B was performed without clothing changes, whereas the matching between Camera views A and C was cross-clothes matching. The results were assessed in terms of the cumulated matching characteristics, specifically, the Rank-1, Rank-10, and Rank-20 matching accuracy.

\section{Experimental Results of Different Settings}

\noindent\textbf{Experimental Setups.} As we described in the main manuscript, we build \textbf{three} kinds of different experiment settings to comprehensively validate the effectiveness of gait biometric for person ReID, and also validate the rationality/superiority of the proposed gait prediction and regularization in our GI-ReID framework: (1) Real Cloth-Changing Image ReID, (2) General Video ReID, (3) Imitated Cloth-Changing Video ReID. In the main manuscripts, we have presented all the results related to the most challenging setting of (1) real cloth-changing image ReID. The rest results about (2)(3) are shown here. Baseline means the model that only ingests RGB images.

\noindent\textbf{2) General Video ReID.} In this setting, we use a general video ReID dataset MARS for experiments. This dataset has no cloth-changing cases. This group of experiments aims to verify two things: 1) gait could benefit ReID even without clothes variations. 2) extracting gait feature from video is easier than that from image, or said, exploiting gait feature in the image-based CC-ReID is more challenging. Since MARS itself contains continuous video frames/clips and human gait masks~\footnote{https://pan.baidu.com/s/16ZrlM1f\_1\_T-eZHmQTTkYg.}, we don't need GSP to additionally predict gait sequence, so we discard it for simplicity.



\noindent\textbf{3) Imitated Cloth-Changing Video ReID.} We still use MARS as dataset to perform experiments in this setting. But the difference is that we \emph{imitate} cloth-changing cases for the images with the same identity through a data augmentation strategy---body-wise color jitter (\ieno, randomly change the brightness, contrast and saturation of the human body region in an person image) for training. This group of experiments aims to show that gait information could alleviate the ReID interference caused by clothes changing. GSP module is also removed in this setting.




\noindent\textbf{Results of General Video ReID.} Table~\ref{tab:tab:GvideoReID} shows the results. We observe that: 1) Thanks to the leverage of gait characteristics through the proposed Gait-Stream (GS), \emph{Baseline + GS (concat)} and \emph{Baseline + GS + SC} outperform \emph{Baseline} by 1.07\%/1.29\% in mAP respectively, which demonstrates that gait information indeed benefits ReID. 2) We find that \emph{Baseline + GS + SC} further outperforms \emph{Baseline + GS (concat)} by 0.22\% in mAP. This result validates the superiority of our gait utilization manner (\ieno, regularization), which makes ReID-Stream not only robust to the gait estimation error, but also computationally efficient (Gait-Stream is not needed in the inference).

\begin{table}[htbp]
  \centering
  \footnotesize
  \vspace{-2mm}
  \caption{Performance (\%) comparison on the general video ReID dataset MARS~\cite{zheng2016mars}. \textbf{GS} refers to Gait-Stream and \textbf{SC} refers to semantics consistency constraints. Note that `concat' means concatenating ReID vector $r$ and gait vector $g$ together for ReID. The backbone is ResNet-50.}
  \vspace{-3mm}
  \setlength{\tabcolsep}{6mm}{
    \begin{tabular}{ccc}
    \toprule
    \multirow{2}[4]{*}{Methods} & \multicolumn{2}{c}{MARS} \\
\cmidrule{2-3}          & mAP   & Rank-1 \\
    \midrule
    Baseline & 79.12 & 87.34  \\
    Baseline + GS (concat)  & 80.19 & 88.16 \\
    Baseline + GS + SC (ours)  & \textbf{80.41} & \textbf{88.32} \\
    \bottomrule
    \end{tabular}}%
    \vspace{-3mm}
  \label{tab:tab:GvideoReID}%
\end{table}%

\noindent\textbf{Results of Imitated Cloth-Changing Video ReID.} To prove that gait indeed could alleviate clothes variation issue, we \emph{imitate} cloth-changing cases for MARS (denoted as CC-MARS). In Table~\ref{tab:tab:CvideoReID}, we observe that 1) Disturbed by the synthetic clothing change, \emph{Baseline} suffers from large degradation, 68.52\% on CC-MARS vs. 79.12\% on raw MARS in mAP. 2) With the assistance of gait, \emph{Baseline+GS (concat)} and \emph{Baseline+GS+SC} improve \emph{Baseline} near 5.0\% in mAP. 3) On CC-MARS, the gait `concat' scheme shows a little superiority than ours. We analyse that's because the `concat' could help ReID more explicitly, especially when meeting changing clothes. But, the `concat' scheme needs maintain Gait-Stream in the inference, leading extra computational cost. 4) As video ReID datasets, it is relatively easy to extract gait features on MARS/CC-MARS. 


\begin{table}[htbp]
  \centering
  \footnotesize
  \caption{Performance (\%) comparison on the imitated (using color jitter) cloth-changing video ReID dataset, termed as CC-MARS. The ReID backbone is ResNet-50.}
  \vspace{-3mm}
  \setlength{\tabcolsep}{6mm}{
    \begin{tabular}{ccc}
    \toprule
    \multirow{2}[4]{*}{Methods} & \multicolumn{2}{c}{CC-MARS} \\
\cmidrule{2-3}          & mAP   & Rank-1 \\
    \midrule
    Baseline & 68.52 & 72.31  \\
    Baseline + GS (concat)  & \textbf{73.46} & \textbf{79.34} \\    
    Baseline + GS + SC (ours)  & 73.13 & 79.15 \\
    \bottomrule
    \end{tabular}}%
  \label{tab:tab:CvideoReID}%
\end{table}%

\section{Comparison with State-of-the-Arts (Complete version)}

\begin{table*}
\centering{}%
\small
\caption{\label{tab:Results-LTCC} 
Performance (\%) comparisons of our GI-ReID and other competitors on the cloth-changing dataset LTCC~\cite{qian2020long}. `Standard' and `Cloth-changing' respectively mean the standard setting and cloth-changing setting as mentioned in our main manuscript. `(Image)' or `(Parsing)' represents that the input data is the person image or the body parsing image. `$\dag$' means the setting that only identities with clothes changing are used for training.}
\vspace{-3mm}
\setlength{\tabcolsep}{2mm}{
\begin{tabular}{@{\extracolsep{\fill}}l|c|c|c|c||c|c|c|c}
\hline 
\multicolumn{1}{c|}{\multirow{2}{*}{Methods}} & \multicolumn{2}{c|}{Standard} & \multicolumn{2}{c||}{Cloth-changing} & \multicolumn{2}{c|}{Standard$^\dag$} & \multicolumn{2}{c}{Cloth-changing$^\dag$}\tabularnewline
\cline{2-9} 
 & Rank-1 & mAP & Rank-1 & mAP & Rank-1 & mAP & Rank-1 & mAP\tabularnewline
\hline 
LOMO \cite{XQDA} + KISSME \cite{crowduser} & 26.57 & 9.11 & 10.75 & 5.25 & 19.47 & 7.37 & 8.32 & 4.37 \tabularnewline
LOMO \cite{XQDA} + XQDA \cite{XQDA} & 25.35 & 9.54 & 10.95 & 5.56 & 22.52 & 8.21 & 10.55 & 4.95 \tabularnewline
LOMO \cite{XQDA} + NullSpace \cite{NullReid} & 34.83 & 11.92 & 16.45 & 6.29 & 27.59 & 9.43 & 13.37 & 5.34 \tabularnewline
\hline 

ResNet-50 (Image) \cite{he2016deep} & 58.82 & 25.98 & 20.08 & 9.02 & 57.20 & 22.82 & 20.68 & 8.38 \tabularnewline

ResNet-50 (Parsing) \cite{he2016deep} & 19.87 & 6.64 & 7.51 & 3.75 & 18.86 & 6.16 & 6.28 & 3.46 \tabularnewline
PCB (Parsing) \cite{sun2018beyond} & 27.38 & 9.16 & 9.33 & 4.50 & 25.96 & 7.77 & 10.54 & 4.04
\tabularnewline
ResNet-50 + Face~\cite{xue2018clothing} & 60.44 & 25.42 & 22.10 & 9.44 & 55.37 & 22.23 & 20.68 & 8.99
\tabularnewline
PCB \cite{sun2018beyond} & 65.11 & 30.60 & 23.52 & 10.03 & 59.22 & 26.61 & 21.93 & 8.81 \tabularnewline
HACNN \cite{li2018harmonious} & 60.24 & 26.71 & 21.59 & 9.25 & 57.12 & 23.48 & 20.81 & 8.27 \tabularnewline
MuDeep \cite{qian2019leader} & 61.86 & 27.52 & 23.53 & 10.23 & 56.99 & 24.10 & 18.66 & 8.76\tabularnewline
Face \cite{xue2018clothing} & 60.44  & 25.42  & 22.10  & 9.44  & 55.37  & 22.23  & 20.68  & 8.99 \tabularnewline


\hline
        
        

        
        
        Baseline (ResNet-50) & 55.14 & 23.21 & 19.58 & 8.10 & 54.27 & 21.98 & 19.14 & 7.74 \\
        
        GI-ReID (ResNet-50, ours) & 63.21 & 29.44 & 23.72 & 10.38 & 61.39 & 27.88 & 22.59 & 9.87  \\
        \hline
        Baseline (OSNet) & 66.07 & 31.18 & 23.43 & 10.56  & 61.22 & 27.41 & 22.97 & 9.74 \\
        
        GI-ReID (OSNet, ours) & \bf 73.59 & \bf36.07 & {28.11} & {13.17} & \bf 66.94 & \bf 33.04 & \bf 26.71 & {12.69} \\
        \hline 
        
        Baseline (LTCC-shape~\cite{qian2020long}) & -- & -- & {26.15} & {12.40} & -- & -- & {25.15} &  {11.67} \\
        
        LTCC-shape + Gait-Stream (ours) & -- & -- & \bf 28.86 & \bf 14.19 & -- & -- & {26.41} &  \bf 13.26 \\

        \hline 
\end{tabular}}
\end{table*}

\begin{table*}[t]
	\setlength{\tabcolsep}{16pt}
	\begin{center}
		\caption{Performance (\%) comparisons of our GI-ReID and other competitors on the cloth-changing dataset PRCC~\cite{yang2019person}. ``RGB'' means the inputs of the model are RGB images; ``Sketch''  means the inputs of the model are contour sketch images}
		\label{tab:PRCC_STOA}
		\resizebox{0.93\textwidth}{!}{%
		\begin{tabular}{l|c|c|c||c|c|c}
			\hline
			\multirow{2}{*}{Methods}
			&\multicolumn{3}{c||}{{Cameras A and C (Cross-Clothes)}}
			&\multicolumn{3}{c}{{Cameras A and B (Same Clothes)}} \\ \cline{2-7}
			                                                                                                     & Rank-1         & Rank-10 & Rank-20 & Rank-1         & Rank-10 & Rank-20 \\\hline
			LBP \cite{ojala1996comparative} + KISSME \cite{koestinger2012large}                                  & 18.71          & 58.09   & 71.40   & 39.03          & 76.18   & 86.91   \\
			HOG \cite{dalal2005histograms} + KISSME \cite{koestinger2012large}                                   & 17.52          & 49.52   & 63.55   & 36.02          & 68.83   & 80.49   \\
			LBP \cite{ojala1996comparative} + HOG \cite{dalal2005histograms} + KISSME \cite{koestinger2012large} & 17.66          & 54.07   & 67.85   & 47.73          & 81.88   & 90.54   \\

			LOMO \cite{liao2015person} + KISSME \cite{koestinger2012large}                                       & 18.55          & 49.81   & 67.27   & 47.40          & 81.42   & 90.38   \\\hline
			LBP \cite{ojala1996comparative} + XQDA \cite{liao2015person}                                         & 18.25          & 52.75   & 61.98   & 40.66          & 77.74   & 87.44   \\
			HOG \cite{dalal2005histograms} + XQDA \cite{liao2015person}                                          & 22.11          & 57.33   & 69.93   & 42.32          & 75.63   & 85.38   \\
			LBP \cite{ojala1996comparative} + HOG \cite{dalal2005histograms} + XQDA \cite{liao2015person}        & 23.71          & 62.04   & 74.49   & 54.16          & 84.11   & 91.21   \\

			LOMO \cite{liao2015person} + XQDA \cite{liao2015person}                                              & 14.53          & 43.63   & 60.34   & 29.41          & 67.24   & 80.52   \\\hline
			Shape \cite{belongie2002shape}                                                               & 11.48          & 38.66   & 53.21   & 23.87          & 68.41   & 76.32   \\
			LNSCT \cite{xie2010extraction}                                                                       & 15.33          & 53.87   & 67.12   & 35.54          & 69.56   & 82.37   \\\hline
			Alexnet \cite{krizhevsky2017imagenet} (RGB)                                                          & 16.33          & 48.01   & 65.87   & 63.28          & 91.70   & 94.73   \\
			VGG16 \cite{simonyan2014very} (RGB)                                                                  & 18.21          & 46.13   & 60.76   & 71.39          & 95.89   & 98.68   \\
			HA-CNN \cite{li2018harmonious} (RGB)                                                                 & 21.81          & 59.47   & 67.45   & 82.45          & 98.12   & 99.04   \\
			PCB \cite{sun2018beyond} (RGB)                                                                       & 22.86          & 61.24   & 78.27   & \textbf{86.88} & {98.79}   & {99.62}   \\\hline
			Alexnet \cite{krizhevsky2017imagenet} (Sketch)                                                       & 14.94          & 57.68   & 75.40   & 38.00          & 82.15   & 91.91   \\
			VGG16 \cite{simonyan2014very} (Sketch)                                                               & 18.79          & 66.01   & 81.27   & 54.00          & 91.33   & 96.73   \\
			HA-CNN \cite{li2018harmonious} (Sketch)                                                              & 20.45          & 63.87   & 79.58   & 58.63          & 90.45   & 95.78   \\
			PCB \cite{sun2018beyond} (Sketch)                                                                    & 22.48          & 61.07   & 77.05   & 57.36          & 92.12   & 96.72   \\
			SketchNet  \cite{zhang2016sketchnet} (Sketch+RGB)                                                    & 17.89          & 43.70   & 58.62   & 64.56          & 95.09   & 97.84   \\
			Face~\cite{wen2016discriminative}                                   & 2.97           & 9.85    & 13.52   & 4.75           & 13.40   & 45.54   \\\hline
			Deformable Conv. \cite{dai2017deformable}                                                            & 25.98          & 71.67   & 85.31   & 61.87          & 92.13   & 97.65   \\
			STN \cite{jaderberg2015spatial}                                                                      & 27.47          & 69.53   & 83.22   & 59.21          & 91.43   & 96.11   \\

            
            RCSANet~\cite{huang2021clothing} & 31.60 & -- & -- & -- & -- & -- \\
            
            \hline
            PRCC-contour~\cite{yang2019person} & {34.38} & {77.30}  & {88.05}  & {64.20} & 92.62   & 96.65\\

            + Gait-Stream (ours)  & {36.19} & {79.93}   & {91,67}  & -- & --   & --  \\
            \hline
            Baseline (ResNet-50) & 22.23 & 61.08 & 76.44 & 75.81 & 97.34 & 98.95\\
            GI-ReID (ResNet-50) & 33.26 & 75.09 & 87.44 & 78.95 & 97.89 & 99.11\\
            \hline
            Baseline (OSNet) & 28.70 & 72.34 & 85.89 & 83.68 & 98.24 & 99.26\\
            GI-ReID (OSNet) & \textbf{37.55} & \textbf{82.25} & \textbf{93.76} &{85.97} &\textbf{98.82} &\textbf{99.72} \\
            \hline

            \hline
		\end{tabular}
		}
	\end{center}
\end{table*}

To save space, we only present the latest approaches in the main manuscripts, and here we show comparisons with more approaches and more evaluation settings on LTCC (Table~\ref{tab:Results-LTCC}) and PRCC datasets (Table~\ref{tab:PRCC_STOA}).

From the comparison results on PRCC that are shown in Table~\ref{tab:PRCC_STOA}, we observe that 1) Although person ReID with no clothing change (\ieno ``Same Clothes'' in the Table~\ref{tab:PRCC_STOA}) is not the purpose in our work, our method GI-ReID can still achieve an accuracy of 85.97\% in Rank-1, which is better than that of all hand-crafted features with metric learning methods and most deep learning methods. 2) When the input images are RGB images without clothing changes, Alexnet~\cite{krizhevsky2017imagenet}, VGG16~\cite{simonyan2014very}, HA-CNN~\cite{li2018harmonious}, and PCB~\cite{sun2018beyond} all achieve good performance, but they have a sharp performance drop when a clothing change occurs, illustrating the challenge of person ReID when a person dresses differently. Therefore, the application of existing person ReID methods is not straightforward in this scenario. In contrast, our GI-ReID that leverages gait information is beneficial to learn the
clothing invariant feature, which makes our method achieve satisfactory performance 37.55\% in Rank-1 even under the cloth-changing scenario. 

\section{Study on Failure Cases (Limitations)}
\begin{figure}[htp]
  \centerline{\includegraphics[width=1.0\linewidth]{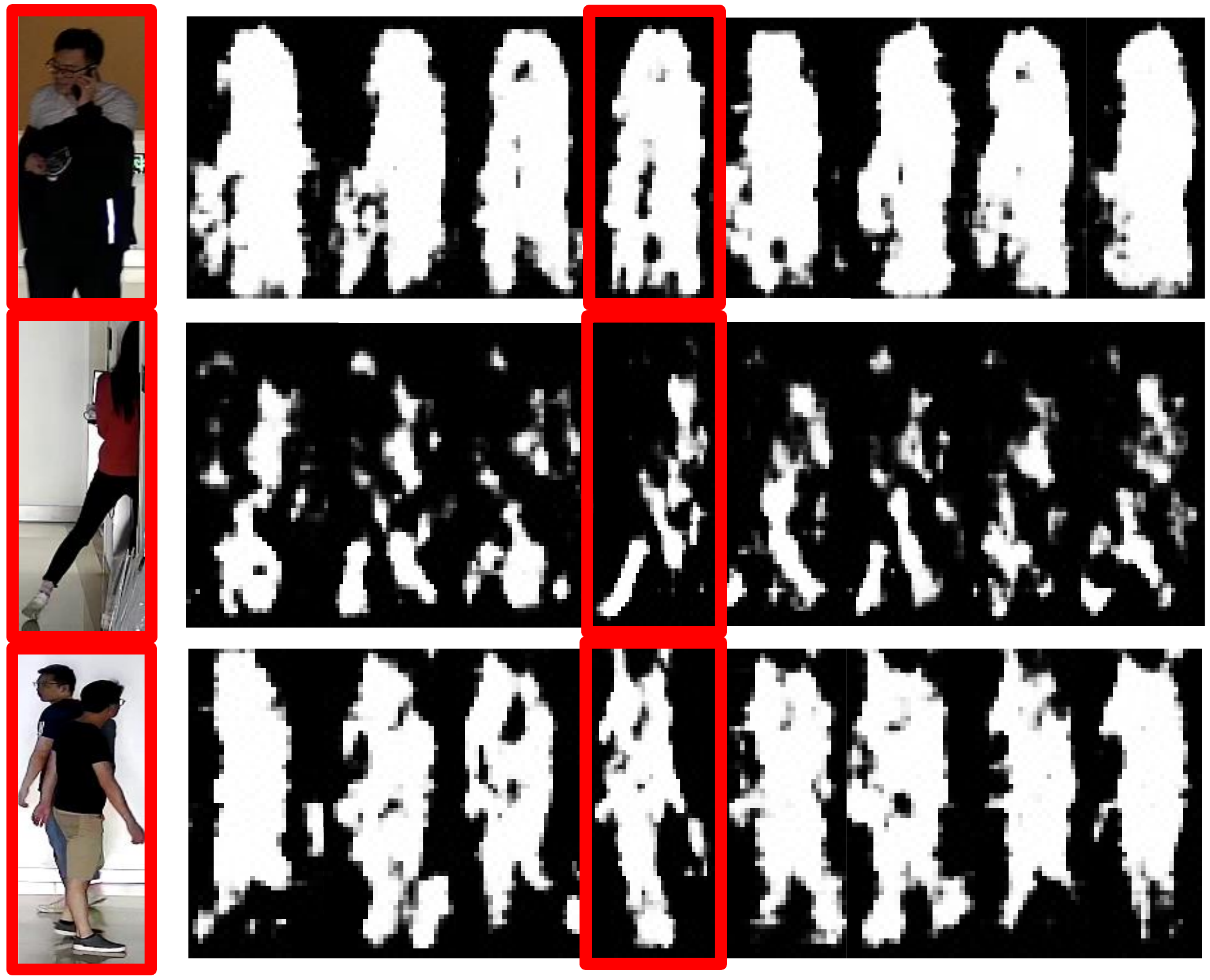}}
  \vspace{-3mm}
  \caption{Failure cases of gait sequence prediction (GSP).}
  \vspace{-5mm}
  \label{fig:supp_gcp_failure}
\end{figure}

As we described in the main manuscript, since the existed large difference on the capture viewpoint and environment between gait and ReID training data, the predicted results of gait sequence prediction (GSP) module are not so accurate when occlusion, partial, multi-person, \etcno, existed in the person images. As shown in Figure~\ref{fig:supp_gcp_failure}, GSP gives unsatisfactory gait prediction results, where large estimation errors exist in the predicted gait frames, which will hurt the ReID performance. That is why we \textbf{indirectly} use gait prediction results in a two-stream knowledge regularization manner, which makes our GI-ReID robust/less sensitive to these failure cases.

\section{Social Impact}

\noindent\textbf{Positive.} In this paper, we propose to utilize human unique gait to address the cloth-changing ReID (CC-ReID) problem from a single image. A novel gait-involved two-stream framework GI-ReID is introduced for image-based CC-ReID. To our best knowledge, this paper is the first attempt to take gait as a regulator with a Gait-Stream (discarded in the inference), to encourage the cloth-agnostic representation learning of image-based ReID-Stream. This is very important for both of academic community and industry, and it is also valuable and meaningful to bridge the gap between the fast-developing ReID algorithms and practical applications.

\noindent\textbf{Negative.} Due to the urgent demand of public safety and
increasing number of surveillance cameras, person ReID is imperative in intelligent surveillance systems with significant research impact and practical importance, but this task also might raise questions about the risk of leaking private information. On the other hand, the data collected from the surveillance equipments or downloaded from the internet may violate the privacy of human beings. Therefore, we appeal and encourage research that understands and mitigates the risks arising from surveillance applications.

{\small
\bibliographystyle{ieee_fullname}
\bibliography{egbib}

\begin{thebibliography}{10}\itemsep=-1pt

\bibitem{belongie2002shape}
Serge Belongie, Jitendra Malik, and Jan Puzicha.
\newblock Shape matching and object recognition using shape contexts.
\newblock {\em IEEE TPAMI}, 24(4):509--522, 2002.

\bibitem{carley2019person}
Cassandra Carley, Ergys Ristani, and Carlo Tomasi.
\newblock Person re-identification from gait using an autocorrelation network.
\newblock In {\em CVPRW}, pages 0--0, 2019.

\bibitem{chao2019gaitset}
Hanqing Chao, Yiwei He, Junping Zhang, and Jianfeng Feng.
\newblock Gaitset: Regarding gait as a set for cross-view gait recognition.
\newblock In {\em AAAI}, volume~33, pages 8126--8133, 2019.

\bibitem{chen2018person}
Di Chen, Shanshan Zhang, Wanli Ouyang, Jian Yang, and Ying Tai.
\newblock Person search via a mask-guided two-stream cnn model.
\newblock In {\em ECCV}, pages 734--750, 2018.

\bibitem{dai2017deformable}
Jifeng Dai, Haozhi Qi, Yuwen Xiong, Yi Li, Guodong Zhang, Han Hu, and Yichen
  Wei.
\newblock Deformable convolutional networks.
\newblock In {\em ICCV}, pages 764--773, 2017.

\bibitem{dalal2005histograms}
Navneet Dalal and Bill Triggs.
\newblock Histograms of oriented gradients for human detection.
\newblock 2005.

\bibitem{doersch2016tutorial}
Carl Doersch.
\newblock Tutorial on variational autoencoders.
\newblock {\em arXiv preprint arXiv:1606.05908}, 2016.

\bibitem{elharrouss2020gait}
Omar Elharrouss, Noor Almaadeed, Somaya Al-Maadeed, and Ahmed Bouridane.
\newblock Gait recognition for person re-identification.
\newblock {\em The Journal of Supercomputing}, pages 1--20, 2020.

\bibitem{Fan_2020_CVPR}
Chao Fan, Yunjie Peng, Chunshui Cao, Xu Liu, Saihui Hou, Jiannan Chi, Yongzhen
  Huang, Qing Li, and Zhiqiang He.
\newblock Gaitpart: Temporal part-based model for gait recognition.
\newblock In {\em CVPR}, June 2020.

\bibitem{fu2019horizontal}
Yang Fu, Yunchao Wei, Yuqian Zhou, Honghui Shi, Gao Huang, Xinchao Wang,
  Zhiqiang Yao, and Thomas Huang.
\newblock Horizontal pyramid matching for person re-identification.
\newblock In {\em AAAI}, volume~33, pages 8295--8302, 2019.

\bibitem{ge2018fd}
Yixiao Ge, Zhuowan Li, Haiyu Zhao, et~al.
\newblock Fd-gan: Pose-guided feature distilling gan for robust person
  re-identification.
\newblock In {\em NeurIPS}, 2018.

\bibitem{gretton2012kernel}
Arthur Gretton, Karsten~M Borgwardt, Malte~J Rasch, Bernhard Sch{\"o}lkopf, and
  Alexander Smola.
\newblock A kernel two-sample test.
\newblock {\em The Journal of Machine Learning Research}, 13(1):723--773, 2012.

\bibitem{guen2020disentangling}
Vincent~Le Guen and Nicolas Thome.
\newblock Disentangling physical dynamics from unknown factors for unsupervised
  video prediction.
\newblock In {\em CVPR}, pages 11474--11484, 2020.

\bibitem{he2016deep}
Kaiming He, Xiangyu Zhang, Shaoqing Ren, et~al.
\newblock Deep residual learning for image recognition.
\newblock In {\em CVPR}, 2016.

\bibitem{he2018deep}
Lingxiao He, Jian Liang, Haiqing Li, and Zhenan Sun.
\newblock Deep spatial feature reconstruction for partial person
  re-identification: Alignment-free approach.
\newblock In {\em CVPR}, 2018.

\bibitem{hermans2017defense}
Alexander Hermans, Lucas Beyer, and Bastian Leibe.
\newblock In defense of the triplet loss for person re-identification.
\newblock {\em arXiv preprint arXiv:1703.07737}, 2017.

\bibitem{hinton2015distilling}
Geoffrey Hinton, Oriol Vinyals, and Jeff Dean.
\newblock Distilling the knowledge in a neural network.
\newblock {\em NeurIPS Workshop}, 2015.

\bibitem{hong2021fine}
Peixian Hong, Tao Wu, Ancong Wu, Xintong Han, and Wei-Shi Zheng.
\newblock Fine-grained shape-appearance mutual learning for cloth-changing
  person re-identification.
\newblock In {\em CVPR}, pages 10513--10522, 2021.

\bibitem{hsieh2018learning}
Jun-Ting Hsieh, Bingbin Liu, De-An Huang, Li~F Fei-Fei, and Juan~Carlos
  Niebles.
\newblock Learning to decompose and disentangle representations for video
  prediction.
\newblock In {\em NeurIPS}, pages 517--526, 2018.

\bibitem{huang2019celebrities}
Yan Huang, Qiang Wu, Jingsong Xu, and Yi Zhong.
\newblock Celebrities-reid: A benchmark for clothes variation in long-term
  person re-identification.
\newblock In {\em IJCNN}, pages 1--8. IEEE, 2019.

\bibitem{huang2021clothing}
Yan Huang, Qiang Wu, JingSong Xu, Yi Zhong, and ZhaoXiang Zhang.
\newblock Clothing status awareness for long-term person re-identification.
\newblock In {\em ICCV}, pages 11895--11904, 2021.

\bibitem{huang2019beyond}
Yan Huang, Jingsong Xu, Qiang Wu, Yi Zhong, Peng Zhang, and Zhaoxiang Zhang.
\newblock Beyond scalar neuron: Adopting vector-neuron capsules for long-term
  person re-identification.
\newblock {\em TCSVT}, 2019.

\bibitem{ioffe2015batch}
Sergey Ioffe and Christian Szegedy.
\newblock Batch normalization: Accelerating deep network training by reducing
  internal covariate shift.
\newblock {\em ICML}, 2015.

\bibitem{jaderberg2015spatial}
Max Jaderberg, Karen Simonyan, Andrew Zisserman, et~al.
\newblock Spatial transformer networks.
\newblock In {\em NeurIPS}, pages 2017--2025, 2015.

\bibitem{jin2020global}
Xin Jin, Cuiling Lan, Wenjun Zeng, and Zhibo Chen.
\newblock Global distance-distributions separation for unsupervised person
  re-identification.
\newblock {\em ECCV}, 2020.

\bibitem{jin2020uncertainty}
Xin Jin, Cuiling Lan, Wenjun Zeng, and Zhibo Chen.
\newblock Uncertainty-aware multi-shot knowledge distillation for image-based
  object re-identification.
\newblock In {\em AAAI}, 2020.

\bibitem{jin2020style}
Xin Jin, Cuiling Lan, Wenjun Zeng, Zhibo Chen, and Li Zhang.
\newblock Style normalization and restitution for generalizable person
  re-identification.
\newblock In {\em CVPR}, pages 3143--3152, 2020.

\bibitem{jin2020semantics}
Xin Jin, Cuiling Lan, Wenjun Zeng, Guoqiang Wei, and Zhibo Chen.
\newblock Semantics-aligned representation learning for person
  re-identification.
\newblock In {\em AAAI}, 2020.

\bibitem{kingma2014adam}
Diederik~P Kingma and Jimmy Ba.
\newblock Adam: A method for stochastic optimization.
\newblock In {\em ICLR}, 2014.

\bibitem{kirillov2020pointrend}
Alexander Kirillov, Yuxin Wu, Kaiming He, and Ross Girshick.
\newblock Pointrend: Image segmentation as rendering.
\newblock In {\em CVPR}, pages 9799--9808, 2020.

\bibitem{crowduser}
Aniket Kittur, Ed~H. Chi, and Bongwon Suh.
\newblock Crowdsourcing user studies with mechanical turk.
\newblock 2008.

\bibitem{koestinger2012large}
Martin Koestinger, Martin Hirzer, Paul Wohlhart, Peter~M Roth, and Horst
  Bischof.
\newblock Large scale metric learning from equivalence constraints.
\newblock In {\em CVPR}, pages 2288--2295. IEEE, 2012.

\bibitem{kourtzi2000activation}
Zoe Kourtzi and Nancy Kanwisher.
\newblock Activation in human mt/mst by static images with implied motion.
\newblock {\em Journal of cognitive neuroscience}, 12(1):48--55, 2000.

\bibitem{krizhevsky2017imagenet}
Alex Krizhevsky, Ilya Sutskever, and Geoffrey~E Hinton.
\newblock Imagenet classification with deep convolutional neural networks.
\newblock {\em Communications of the ACM}, 60(6):84--90, 2017.

\bibitem{li2018harmonious}
Wei Li, Xiatian Zhu, and Shaogang Gong.
\newblock Harmonious attention network for person re-identification.
\newblock In {\em CVPR}, 2018.

\bibitem{li2020gait}
Xiang Li, Yasushi Makihara, Chi Xu, Yasushi Yagi, and Mingwu Ren.
\newblock Gait recognition via semi-supervised disentangled representation
  learning to identity and covariate features.
\newblock In {\em CVPR}, pages 13309--13319, 2020.

\bibitem{li2020end}
Xiang Li, Yasushi Makihara, Chi Xu, Yasushi Yagi, Shiqi Yu, and Mingwu Ren.
\newblock End-to-end model-based gait recognition.
\newblock In {\em ACCV}, 2020.

\bibitem{li2020learning}
Yu-Jhe Li, Zhengyi Luo, Xinshuo Weng, and Kris~M Kitani.
\newblock Learning shape representations for clothing variations in person
  re-identification.
\newblock {\em arXiv preprint arXiv:2003.07340}, 2020.

\bibitem{liao2020model}
Rijun Liao, Shiqi Yu, Weizhi An, and Yongzhen Huang.
\newblock A model-based gait recognition method with body pose and human prior
  knowledge.
\newblock {\em Pattern Recognition}, 98:107069, 2020.

\bibitem{XQDA}
S. Liao, Y. Hu, X. Zhu, and S.~Z. Li.
\newblock Person re-identification by local maximal occurrence representation
  and metric learning.
\newblock In {\em CVPR}, 2015.

\bibitem{liao2015person}
Shengcai Liao, Yang Hu, Xiangyu Zhu, and Stan~Z Li.
\newblock Person re-identification by local maximal occurrence representation
  and metric learning.
\newblock In {\em CVPR}, 2015.

\bibitem{liu2019deep}
Yu-Lun Liu, Yi-Tung Liao, Yen-Yu Lin, and Yung-Yu Chuang.
\newblock Deep video frame interpolation using cyclic frame generation.
\newblock In {\em AAAI}, volume~33, pages 8794--8802, 2019.

\bibitem{liu2015enhancing}
Zheng Liu, Zhaoxiang Zhang, Qiang Wu, and Yunhong Wang.
\newblock Enhancing person re-identification by integrating gait biometric.
\newblock {\em Neurocomputing}, 168:1144--1156, 2015.

\bibitem{makihara2017joint}
Yasushi Makihara, Atsuyuki Suzuki, Daigo Muramatsu, Xiang Li, and Yasushi Yagi.
\newblock Joint intensity and spatial metric learning for robust gait
  recognition.
\newblock In {\em CVPR}, pages 5705--5715, 2017.

\bibitem{meyer2015phase}
Simone Meyer, Oliver Wang, Henning Zimmer, Max Grosse, and Alexander
  Sorkine-Hornung.
\newblock Phase-based frame interpolation for video.
\newblock In {\em CVPR}, pages 1410--1418, 2015.

\bibitem{miao2019pose}
Jiaxu Miao, Yu Wu, Ping Liu, Yuhang Ding, and Yi Yang.
\newblock Pose-guided feature alignment for occluded person re-identification.
\newblock In {\em ICCV}, pages 542--551, 2019.

\bibitem{muramatsu2014gait}
Daigo Muramatsu, Akira Shiraishi, Yasushi Makihara, Md~Zasim Uddin, and Yasushi
  Yagi.
\newblock Gait-based person recognition using arbitrary view transformation
  model.
\newblock {\em TIP}, 24(1):140--154, 2014.

\bibitem{nair2010rectified}
Vinod Nair and Geoffrey~E Hinton.
\newblock Rectified linear units improve restricted boltzmann machines.
\newblock In {\em ICML}, 2010.

\bibitem{niklaus2018context}
Simon Niklaus and Feng Liu.
\newblock Context-aware synthesis for video frame interpolation.
\newblock In {\em CVPR}, pages 1701--1710, 2018.

\bibitem{nixon2009model}
Mark Nixon et~al.
\newblock Model-based gait recognition.
\newblock 2009.

\bibitem{ojala1996comparative}
Timo Ojala, Matti Pietik{\"a}inen, and David Harwood.
\newblock A comparative study of texture measures with classification based on
  featured distributions.
\newblock {\em Pattern recognition}, 29(1):51--59, 1996.

\bibitem{qian2018pose}
Xuelin Qian, Yanwei Fu, Wenxuan Wang, et~al.
\newblock Pose-normalized image generation for person re-identification.
\newblock In {\em ECCV}, 2018.

\bibitem{qian2019leader}
Xuelin Qian, Yanwei Fu, Tao Xiang, Yu-Gang Jiang, and Xiangyang Xue.
\newblock Leader-based multi-scale attention deep architecture for person
  re-identification.
\newblock {\em TPAMI}, 2019.

\bibitem{qian2020long}
Xuelin Qian, Wenxuan Wang, Li Zhang, Fangrui Zhu, Yanwei Fu, Tao Xiang, Yu-Gang
  Jiang, and Xiangyang Xue.
\newblock Long-term cloth-changing person re-identification.
\newblock {\em WACV}, 2020.

\bibitem{sabour2017dynamic}
Sara Sabour, Nicholas Frosst, and Geoffrey~E Hinton.
\newblock Dynamic routing between capsules.
\newblock In {\em NeurIPS}, pages 3856--3866, 2017.

\bibitem{simonyan2014very}
Karen Simonyan and Andrew Zisserman.
\newblock Very deep convolutional networks for large-scale image recognition.
\newblock {\em arXiv preprint arXiv:1409.1556}, 2014.

\bibitem{song2018mask}
Chunfeng Song, Yan Huang, Wanli Ouyang, and Liang Wang.
\newblock Mask-guided contrastive attention model for person re-identification.
\newblock In {\em CVPR}, 2018.

\bibitem{su2017pose}
Chi Su, Jianing Li, Shiliang Zhang, et~al.
\newblock Pose-driven deep convolutional model for person re-identification.
\newblock In {\em ICCV}, 2017.

\bibitem{sun2018dissecting}
Xiaoxiao Sun and Liang Zheng.
\newblock Dissecting person re-identification from the viewpoint of viewpoint.
\newblock {\em arXiv preprint arXiv:1812.02162}, 2018.

\bibitem{sun2019dissecting}
Xiaoxiao Sun and Liang Zheng.
\newblock Dissecting person re-identification from the viewpoint of viewpoint.
\newblock In {\em CVPR}, pages 608--617, 2019.

\bibitem{sun2018beyond}
Yifan Sun, Liang Zheng, Yi Yang, Qi Tian, and Shengjin Wang.
\newblock Beyond part models: Person retrieval with refined part pooling (and a
  strong convolutional baseline).
\newblock In {\em ECCV}, pages 480--496, 2018.

\bibitem{takemura2018multi}
Noriko Takemura, Yasushi Makihara, Daigo Muramatsu, Tomio Echigo, and Yasushi
  Yagi.
\newblock Multi-view large population gait dataset and its performance
  evaluation for cross-view gait recognition.
\newblock {\em IPSJ Transactions on Computer Vision and Applications}, 10(1):4,
  2018.

\bibitem{wan2020person}
Fangbin Wan, Yang Wu, Xuelin Qian, Yixiong Chen, and Yanwei Fu.
\newblock When person re-identification meets changing clothes.
\newblock In {\em CVPRW}, pages 830--831, 2020.

\bibitem{wang2018learning}
Guanshuo Wang, Yufeng Yuan, Xiong Chen, et~al.
\newblock Learning discriminative features with multiple granularities for
  person re-identification.
\newblock In {\em ACM MM}, pages 274--282, 2018.

\bibitem{wen2016discriminative}
Yandong Wen, Kaipeng Zhang, Zhifeng Li, and Yu Qiao.
\newblock A discriminative feature learning approach for deep face recognition.
\newblock In {\em ECCV}, pages 499--515. Springer, 2016.

\bibitem{xie2010extraction}
Xiaohua Xie, Jianhuang Lai, and Wei-Shi Zheng.
\newblock Extraction of illumination invariant facial features from a single
  image using nonsubsampled contourlet transform.
\newblock {\em Pattern Recognition}, 43(12):4177--4189, 2010.

\bibitem{xugait2020gait}
Chi Xu, Yasushi Makihara, Xiang Li, Yasushi Yagi, and Jianfeng Lu.
\newblock Gait recognition from a single image using a phase-aware gait cycle
  reconstruction network. eccv 2020.
\newblock 2020.

\bibitem{xue2018clothing}
Jia Xue, Zibo Meng, Karthik Katipally, Haibo Wang, and Kees van Zon.
\newblock Clothing change aware person identification.
\newblock In {\em CVPR Workshops}, pages 2112--2120, 2018.

\bibitem{yang2019person}
Qize Yang, Ancong Wu, and Wei-Shi Zheng.
\newblock Person re-identification by contour sketch under moderate clothing
  change.
\newblock {\em TPAMI}, 2019.

\bibitem{ye2019student}
Jingwen Ye, Yixin Ji, Xinchao Wang, Kairi Ou, Dapeng Tao, and Mingli Song.
\newblock Student becoming the master: Knowledge amalgamation for joint scene
  parsing, depth estimation, and more.
\newblock In {\em CVPR}, pages 2829--2838, 2019.

\bibitem{yu2020cocas}
Shijie Yu, Shihua Li, Dapeng Chen, Rui Zhao, Junjie Yan, and Yu Qiao.
\newblock Cocas: A large-scale clothes changing person dataset for
  re-identification.
\newblock In {\em CVPR}, pages 3400--3409, 2020.

\bibitem{zhang2016sketchnet}
Hua Zhang, Si Liu, Changqing Zhang, Wenqi Ren, Rui Wang, and Xiaochun Cao.
\newblock Sketchnet: Sketch classification with web images.
\newblock In {\em CVPR}, pages 1105--1113, 2016.

\bibitem{NullReid}
Li Zhang, Tao Xiang, and Shaogang Gong.
\newblock Learning a discriminative null space for person re-identificatio.
\newblock In {\em CVPR}, 2016.

\bibitem{zhang2016learning}
Li Zhang, Tao Xiang, and Shaogang Gong.
\newblock Learning a discriminative null space for person re-identification.
\newblock In {\em CVPR}, 2016.

\bibitem{zhang2018long}
Peng Zhang, Qiang Wu, Jingsong Xu, and Jian Zhang.
\newblock Long-term person re-identification using true motion from videos.
\newblock In {\em WACV}, pages 494--502. IEEE, 2018.

\bibitem{zhang2018deep}
Ying Zhang, Tao Xiang, Timothy~M Hospedales, and Huchuan Lu.
\newblock Deep mutual learning.
\newblock In {\em CVPR}, pages 4320--4328, 2018.

\bibitem{zhang2019DSA}
Zhizheng Zhang, Cuiling Lan, Wenjun Zeng, et~al.
\newblock Densely semantically aligned person re-identification.
\newblock In {\em CVPR}, 2019.

\bibitem{zhao2017spindle}
Haiyu Zhao, Maoqing Tian, Shuyang Sun, et~al.
\newblock Spindle net: Person re-identification with human body region guided
  feature decomposition and fusion.
\newblock In {\em CVPR}, 2017.

\bibitem{zheng2016mars}
Liang Zheng, Zhi Bie, Yifan Sun, Jingdong Wang, Chi Su, Shengjin Wang, and Qi
  Tian.
\newblock Mars: A video benchmark for large-scale person re-identification.
\newblock In {\em ECCV}, pages 868--884. Springer, 2016.

\bibitem{zheng2015partial}
Wei-Shi Zheng, Xiang Li, Tao Xiang, Shengcai Liao, Jianhuang Lai, and Shaogang
  Gong.
\newblock Partial person re-identification.
\newblock In {\em ICCV}, 2015.

\bibitem{zhou2019omni}
Kaiyang Zhou, Yongxin Yang, Andrea Cavallaro, et~al.
\newblock Omni-scale feature learning for person re-identification.
\newblock {\em ICCV}, 2019.

\bibitem{zhuo2018occluded}
Jiaxuan Zhuo, Zeyu Chen, Jianhuang Lai, and Guangcong Wang.
\newblock Occluded person re-identification.
\newblock In {\em ICME}, pages 1--6. IEEE, 2018.

\end{thebibliography}
}

\end{document}